\documentclass{article} 
\usepackage{style/arxiv}

\usepackage{url}
\usepackage{import}
\usepackage{multirow}
\usepackage{subcaption}
\usepackage{graphicx}
\usepackage{amsmath}
\usepackage[table]{xcolor}
\usepackage{soul}
\usepackage{tabularx}
\usepackage{booktabs}
\usepackage{amssymb}
\usepackage{natbib}
\usepackage{xspace}
\usepackage{xcolor}
\usepackage{tabularx}

\usepackage[pagebackref=true,breaklinks=true,colorlinks,bookmarks=false]{hyperref}

\newcommand{\EER}{\mathrm{EER}}
\newcommand{\Neff}{N_{\mathrm{eff}}}
\newcommand{\atanh}{\operatorname{atanh}}
\newcommand{\Var}{\operatorname{Var}}
\newcommand{\E}{\operatorname{E}}
\newcommand{\tr}{\operatorname{tr}}
\newcommand{\ICC}{\mathrm{ICC}}
\newcommand{\KCC}{\mathrm{KCC}}
\newcommand{\dprime}{d'}
\newcommand{\inwords}[1]{\begin{quote}\small\emph{In words:} #1\end{quote}}

\begin{document}

\title{How Many Persistent Features Does a Biometric Need? A Law Linking Equal Error Rate to Feature Count and Persistence}
\renewcommand{\shorttitle}{A Law Linking Equal Error Rate to Feature Count and Persistence}

\author{
    Oleg Komogortsev, Mehedi Hasan Raju\\
    Texas State University, San Marcos, Texas, USA\\
    {\small\texttt{ok@txstate.edu, m.raju@txstate.edu}}
}
\date{}

\maketitle
\begin{abstract}

A biometric designer controls the number of features and their temporal persistence. Earlier empirical work linked the equal error rate (EER) to persistence and to the required number of features by straight lines. We derive the law behind these regularities. For $N$ independent standardized Gaussian features with test--retest correlation $\rho$, scored by cosine similarity, Fisher's $z$-transform turns the genuine and impostor scores into two normals, giving $\EER=\Phi(-\tfrac12\sqrt{N}\,\atanh\rho)$ and, inversely, $N=(2q/\atanh\rho)^2$ with $q=\Phi^{-1}(1-\EER)$; intercorrelated features enter through an effective count $\Neff$. The central extension lets persistence vary across comparisons with spread $\tau$ on Fisher's scale: $\EER=\Phi(-m/(\sqrt{\tau^2+v}+\sqrt v))$ with $v=1/(\Neff-2)$, which for a normal spread has a floor $\Phi(-m/\tau)$ that no number of features can lower. With no fitted parameter the simple law reproduces published simulations and the EER--ICC line, and predicts the KCC--EER relation of eye-movement embeddings on GazeBase ($R^2=0.88$), though less well on GazeBaseVR ($0.63$). Read from the genuine and impostor scores, the general form matches the EER of eye movements, keystroke dynamics, the ECG and face embeddings over 101 conditions with $R^2=0.98$ and a median error of $0.6$ points, against $0.87$ and $1.9$ points for the simple law. The 61 eye-movement, keystroke and ECG conditions check consistency on the same scores; the 40 face conditions use held-out subjects, with the same accuracy. The product $\tau\sqrt{\Neff-2}$ gives a design rule: adding or decorrelating features pays fully while it is well below one and little once it exceeds about three, where only higher or more uniform persistence helps.

\end{abstract}


\section{Introduction}
Anyone who has designed a biometric system has asked two questions. How good must my features be? And how many of them do I need? The community has a large empirical literature on the second question for particular modalities (for eye movements, see \cite{Rigas2017BioEye}) and a much thinner one on the first, mostly because ``how good'' is hard to define independently of a classifier. Our group has argued over the past decade that a classifier-independent answer exists: the value of a feature for authentication is governed by its \emph{temporal persistence}, the extent to which a subject's value of the feature is the same when measured again on a later occasion \cite{Friedman2017PLoS,Friedman2020Sensors}. Persistence can be measured by the intraclass correlation coefficient (ICC) \cite{Shrout1979} for approximately normal features or by Kendall's coefficient of concordance (KCC) \cite{Kendall1939} when normality cannot be assumed \cite{Raju2025IJCB,Raju2026TBIOM}.

Three quantitative regularities have emerged from that programme, each obtained by fitting a line to observed performance:
\begin{enumerate}
  \item Across 29 feature subsets drawn from twelve datasets (eleven real ones in four modalities---eye movements, gait, faces, brain structure---and one synthetic), the equal error rate of a PCA-plus-cosine-similarity pipeline fell on the line $\EER\approx46.7-47.7\times\text{median ICC}$ with a coefficient of determination of $0.92$ \cite{Friedman2017PLoS}. The intercept and slope are close enough to $50$ that a ``biometric law'' $\EER=50-50\,\ICC$ was proposed.
  \item For synthetic, mutually independent, normally distributed features, the number of features needed to reach a target EER obeys $\log_{10}N\approx-2.0\,\ICC+c$, with a target-dependent intercept $c$ \cite{Friedman2019Linear}.
  \item For 128-dimensional embeddings produced by a deep eye-movement biometric network \cite{Lohr2022EKYT}, sixty-one signal-quality manipulations produced EER and KCC values on the line $\EER\approx69.7-76.0\,\KCC$ with a reported $R^2=0.92$, while the intercorrelation among embedding dimensions stayed near $0.19$ throughout \cite{Raju2026TBIOM}.
\end{enumerate}
A fourth observation explains the mechanism qualitatively: raising the persistence of a feature set raises the median of the genuine similarity-score distribution and narrows it, while the impostor distribution is unaffected \cite{Friedman2020Sensors}.

These findings are consistent with one another, but they are descriptions, not a theory. Each line was fitted to data; none says why the slope is what it is, whether the line should be straight at all, what happens outside the sampled range, or how the number of features and their intercorrelation enter. The straight line, in particular, cannot be general: a system with many independent features reaches $\EER\approx0$ long before $\ICC=1$, and a system with very few reaches it only at $\ICC=1$. What is needed---in the words of the late Lee Friedman, who initiated this line of work---is ``a mathematical model of how increased ICC leads to an elevated mean genuine similarity score, and how this relationship is affected by the intercorrelation of the features.''

This paper supplies that model. Its core is short. When features are standardized and independent, the cosine similarity between a subject's enrollment vector and a probe vector is a Pearson correlation coefficient computed from $N$ paired numbers; for a genuine comparison the pairs are correlated with the test--retest correlation $\rho$ of the features, for an impostor comparison they are uncorrelated. Everything about the error rates then follows from the classical sampling theory of the correlation coefficient \cite{Fisher1915,Fisher1921,Hotelling1953}. The main result is that
\begin{equation}
  \EER=\Phi\!\left(-\tfrac12\sqrt{N}\,\atanh\rho\right),
  \label{eq:law-intro}
\end{equation}
where $\Phi$ is the standard normal distribution function. Equation~\eqref{eq:law-intro} assumes that every genuine comparison has the same persistence. In practice persistence varies from one comparison to the next, between people and between pairs of recordings, and allowing for that spread gives a general form with two terms (Section~\ref{sec:subjects}): a sampling term that shrinks as the number of effective features grows, and a spread term that does not. The question for any biometric is then which of the two limits it: the number of effective features, or the spread of persistence. The remainder of the paper develops the law, extends it to the general form, and tests both on real data.

Analytical models of verification error have a long history. Golfarelli et al.\ \cite{Golfarelli1997} derived false-acceptance and false-rejection rates from multinormal feature distributions under an optimal Bayesian decision rule. Daugman \cite{Daugman1993,Daugman2006} read a number of degrees of freedom off the spread of impostor scores of binary iris codes, an argument that the estimator \eqref{eq:neff-imp} carries over to cosine scores. For two Gaussian score distributions, Poh and Bengio \cite{Poh2005} gave the EER as a function of the F-ratio $(\mu_g-\mu_i)/(\sigma_g+\sigma_i)$, which is the two-normal form used here. These results take the feature or score distributions as given. Here those distributions are derived from the persistence and the number of features, so the separation of the scores acquires a closed form in $\rho$, $N$ and the spread of persistence.

Three neighbouring lines of work are also relevant. The participation ratio that defines $\Neff$ is established in climate science as the effective number of spatial degrees of freedom \cite{Bretherton1999}. Capacity analyses ask how many identities a representation can resolve at a given false-acceptance rate \cite{Gong2019Capacity}, not how the error follows from persistence. And differences in performance between subjects have been described as a biometric menagerie \cite{Doddington1998,Yager2010}, while differences between samples are the subject of quality measures \cite{Grother2007}; the spread $\tau$ of Section~\ref{sec:subjects} gives both a common scale and states their effect on the EER.

\subsection*{Contributions}
\begin{enumerate}
  \item An exact expression for the EER of $N$ independent Gaussian features of persistence $\rho$ under cosine scoring, together with the closed-form law \eqref{eq:law-intro} as its large-$N$ limit and a simple finite-$N$ correction (Section~\ref{sec:derivation}).
  \item The law in KCC units, using the exact identity between two-rater KCC and Spearman's rank correlation (Section~\ref{sec:kcc}).
  \item Extensions to features of unequal persistence, to intercorrelated features via an effective number of independent features, to decorrelated (whitened) features via the \emph{canonical persistence spectrum}, to session-level shifts and rescaling, and to averaged enrollment (Section~\ref{sec:extensions}).
  \item A general form of the law for persistence that varies across comparisons, between people or between recordings: a closed form with a sampling term and a spread term, a mixture form for heavy-tailed spreads, a floor that no number of features can lower, and an estimator of the spread from two sessions of genuine and impostor scores (Section~\ref{sec:subjects}).
  \item A simulation study reproducing, with no fitted parameter, the feature-count table of \cite{Friedman2019Linear}, the score-distribution findings of \cite{Friedman2020Sensors}, and the EER surface over $N=2$--$256$ and $\ICC=0.05$--$0.95$ (Section~\ref{sec:simulation}).
  \item Tests of the general form on four modalities---eye-movement embeddings, keystroke dynamics, the ECG, and face embeddings on held-out subjects---which it matches over 101 conditions with $R^2=0.98$ and no fitted parameter (a consistency check on the same scores for the first three modalities, a held-out test for faces), and which span its range: keystroke dynamics limited by the number of effective features, the ECG mostly by the spread of persistence and faces close to that limit, eye movements by both (Sections~\ref{sec:ekyt-scores}--\ref{sec:modalities}).
  \item A reinterpretation of the published straight lines: the EER--ICC line of \cite{Friedman2017PLoS} is the law's curve for about eight effective independent features, and the KCC--EER relation of \cite{Raju2026TBIOM} is predicted on GazeBase by the law with $\Neff\approx16$ computed a priori from the embeddings' intercorrelation, a value confirmed by the impostor score distributions of the embeddings themselves; on GazeBaseVR the fit is weaker because $\Neff$ itself moves (Section~\ref{sec:realdata}).
  \item Design rules that follow: how many persistent features are needed for a target EER, why intercorrelation caps the benefit of adding dimensions, which lever (persistence, its uniformity, decorrelation, dimension) is worth pulling in each regime, and which form of the law to use for given score distributions (Section~\ref{sec:design}).
\end{enumerate}
Section~\ref{sec:limitations} states the assumptions, their consequences, and a staged protocol for testing the law directly on raw embeddings, of which the first three stages are carried out here. The Supplementary Material lists the code (Section~S20) and gives two short derivations and the numerical method for the exact finite-$N$ form (Sections~S1--S3), together with further tables and analyses and the full validation protocol (Sections~S4--S19).

\section{Background}
\label{sec:background}

\subsection{Verification, Score Distributions, and the EER}
In a verification (authentication) system a probe sample is compared with an enrolled template and a similarity score $s$ is produced. Comparisons of a subject with his or her own template are \emph{genuine}; comparisons with another subject's template are \emph{impostor}. A threshold $t$ accepts the claim when $s\ge t$. The false rejection rate $\mathrm{FRR}(t)=P(s_{\mathrm{gen}}<t)$ and the false acceptance rate $\mathrm{FAR}(t)=P(s_{\mathrm{imp}}\ge t)$ trade off as $t$ moves; the equal error rate is their common value at the threshold where they coincide \cite{Jain2004}.

For two normal score distributions with equal variance $\sigma^2$ and means separated by $\Delta$, the EER has a closed form. Define the sensitivity index of signal-detection theory, $\dprime=\Delta/\sigma$ \cite{Green1966}. The threshold at which the two error rates are equal lies midway between the means, and
\begin{equation}
  \EER=\Phi\!\left(-\dprime/2\right).
  \label{eq:eer-dprime}
\end{equation}
Our derivation reduces the biometric problem to exactly this situation.

\subsection{Three Measures of Temporal Persistence}
\label{sec:indices}
Let $x_{im}$ be the value of one feature for subject $i$ on session (occasion) $m\in\{1,2\}$. Three indices of how well session~1 predicts session~2 are in use.

\subsubsection{\texorpdfstring{Pearson test--retest correlation $r$}{Pearson test-retest correlation r}} the ordinary correlation between $\{x_{i1}\}$ and $\{x_{i2}\}$ across subjects. It measures \emph{consistency}: a constant shift or rescaling of all session-2 values leaves it unchanged.

\subsubsection{Intraclass correlation coefficient} the two-way random-effects, single-measure, absolute-agreement ICC, written $\ICC(2,1)$ by Shrout and Fleiss \cite{Shrout1979}, is the index used in \cite{Friedman2017PLoS}. Under the model $x_{im}=\mu+a_i+b_m+e_{im}$ with subject effect $a_i\sim\mathcal N(0,\sigma_a^2)$, session effect $b_m\sim\mathcal N(0,\sigma_b^2)$ and error $e_{im}\sim\mathcal N(0,\sigma_e^2)$,
\begin{equation}
  \ICC(2,1)=\frac{\sigma_a^2}{\sigma_a^2+\sigma_b^2+\sigma_e^2},
\end{equation}
estimated from the mean squares of a two-way ANOVA. It measures \emph{absolute agreement}: a session effect ($\sigma_b^2>0$) lowers it even when $r$ is unchanged. When there is no session effect, $\ICC(2,1)$ and $r$ estimate the same quantity, $\rho=\sigma_a^2/(\sigma_a^2+\sigma_e^2)$.

\subsubsection{Kendall's coefficient of concordance} $W$ measures the agreement among $m$ rankings of $n$ objects \cite{Kendall1939}. With the two sessions as the ``raters'' and the subjects as the objects, $W$ is a rank-based persistence index that requires no distributional assumption; it was adopted in \cite{Raju2026TBIOM} because 30 of 128 embedding dimensions were non-normal. For two raters without ties, $W$ is an exact function of Spearman's rank correlation $r_s$ between the two sessions,
\begin{equation}
  W=\frac{1+r_s}{2},
  \label{eq:kcc-spearman}
\end{equation}
so that $W=0.5$ means no persistence and $W=1$ perfect persistence (Supplementary Material, Section~S1). For jointly normal data Spearman's and Pearson's coefficients are related by Pearson's classical formula $r_s=(6/\pi)\arcsin(\rho/2)$ \cite{Pearson1907,Kruskal1958}, which differs from $\rho$ by at most $0.018$. Table~\ref{tab:indices} summarizes the relations; the practical reading is $\rho\approx2W-1$.

\begin{table}[t]
\centering
\caption{Persistence indices for two sessions, their relation to the test--retest correlation $\rho$ of a normal feature without session effect, and the per-dimension change of one session to which each is invariant.}
\label{tab:indices}
\footnotesize
\setlength{\tabcolsep}{4pt}
\begin{tabular}{@{}lll@{}}
\toprule
Index & Relation to $\rho$ & Invariant to \\
\midrule
Pearson $r$ & $r=\rho$ & linear change \\
$\ICC(2,1)$ & $=\rho$ if no session effect & nothing \\
Spearman $r_s$ & $r_s=\tfrac{6}{\pi}\arcsin(\rho/2)\approx\rho$ & monotone change \\
Kendall $W$ & $W=\tfrac12(1+r_s)\approx\tfrac12(1+\rho)$ & monotone change \\
\bottomrule
\end{tabular}
\end{table}

\subsection{Effective Number of Independent Features}
Correlated variables carry less than one unit of information each. In genetics the ``effective number of independent tests'' among correlated markers is estimated from the eigenvalues of their correlation matrix \cite{Cheverud2001,Nyholt2004}; in iris recognition Daugman estimated the number of independent bits in an IrisCode from the variance of impostor Hamming distances, finding 173 degrees of freedom in a 2048-bit code in the original study \cite{Daugman1993} and about 250 in a later, much larger one \cite{Daugman2006}. We use the same idea: a correlated feature set behaves, for the purpose of the law, like a smaller set of independent features, and the impostor score distribution reveals how many (Section~\ref{sec:neff}).

\section{The Law for Independent Features}
\label{sec:derivation}

\subsection{Generative Model}
\label{sec:model}
Consider $N$ features measured on $n$ subjects in two sessions. Write the value of feature $k$ for subject $i$ in session $m$ as
\begin{equation}
  x_{ikm}=a_{ik}+e_{ikm},\qquad a_{ik}\sim\mathcal N(0,\rho),\quad e_{ikm}\sim\mathcal N(0,1-\rho),
  \label{eq:model}
\end{equation}
all terms independent. The trait part $a_{ik}$ is fixed for the subject and shared by both sessions; the noise $e_{ikm}$ is drawn afresh in each session. Each feature has unit variance, and the fraction of that variance carried by the trait, $\rho$, is both the population ICC and the Pearson test--retest correlation of the feature. Model \eqref{eq:model} is exactly the generator used in the earlier synthetic-feature studies \cite{Friedman2020Sensors,Friedman2020Gallery,Friedman2019Linear}, which add $\mathcal N\big(0,(1-\rho)/\rho\big)$ noise to a shared standard-normal draw and then $z$-score: the ratio of trait to total variance is the same.

\inwords{Each feature is a fixed personal value plus a fresh measurement error. The ICC is the share of the variance that is personal.}

\subsection{Cosine Similarity Is a Correlation Coefficient}
\begin{figure*}[t]
\centering
\includegraphics[width=\textwidth]{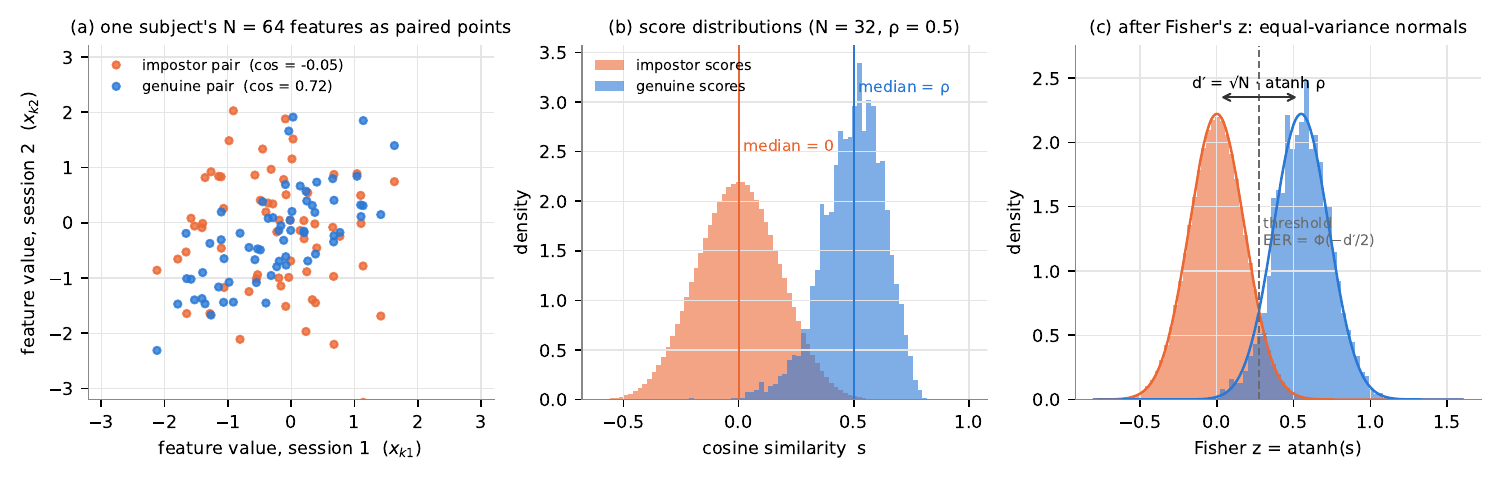}
\caption{The model in pictures. (a) A subject's $N=64$ standardized features plotted as pairs (session-1 value, session-2 value). For a genuine comparison the pairs are correlated with the feature persistence $\rho$ (here $0.7$); for an impostor comparison they are uncorrelated. The cosine similarity of the two feature vectors is the correlation of the plotted cloud about the origin. (b) Genuine and impostor cosine-score distributions for $N=32$ features with $\rho=0.5$ (2,500 simulated subjects): the genuine median is $\rho$, the impostor median is $0$. (c) The same scores after Fisher's transform $z=\atanh s$: two normal distributions with equal variance $1/N$, separated by $\dprime=\sqrt N\,\atanh\rho$; the EER is $\Phi(-\dprime/2)$.}
\label{fig:concept}
\end{figure*}

The enrollment template of subject $i$ is the vector $\mathbf x_{i1}=(x_{i11},\dots,x_{iN1})$ of session-1 values; the probe is $\mathbf x_{j2}$, the session-2 vector of subject $j$ ($j=i$ for a genuine comparison). The cosine similarity is
\begin{equation}
  s=\frac{\mathbf x_{i1}\cdot\mathbf x_{j2}}{\|\mathbf x_{i1}\|\,\|\mathbf x_{j2}\|}
   =\frac{\sum_k x_{ik1}x_{jk2}}{\sqrt{\sum_k x_{ik1}^2}\sqrt{\sum_k x_{jk2}^2}} .
  \label{eq:cosine}
\end{equation}
Read the right-hand side as a statistic of the $N$ pairs $(x_{ik1},x_{jk2})$, $k=1,\dots,N$: it is the Pearson correlation coefficient of those pairs computed about zero rather than about the sample means (Fig.~\ref{fig:concept}a). Because the features are standardized, the pairs have zero population means; because they are independent across $k$, the pairs are independent draws from one bivariate normal distribution. For a genuine comparison that distribution has correlation $\rho$ (both members share $a_{ik}$); for an impostor comparison it has correlation $0$.

The distribution of a sample correlation coefficient computed from independent bivariate-normal pairs is a classical result \cite{Fisher1915}. When $N$ pairs are correlated about zero, the statistic has the same distribution as an ordinary (mean-corrected) sample correlation with $\nu=N$ degrees of freedom, i.e.\ as one computed from $N+1$ pairs. Everything about biometric performance now follows from that distribution.

\subsection{Fisher's Transform and the Law}
Fisher \cite{Fisher1921} showed that $z=\atanh r$ has an approximately normal distribution with mean $\atanh\rho$ and a variance that, to leading order, is $1/\nu$ \emph{whatever the value of $\rho$}. Applied to our two families of scores this gives
\begin{equation}
  z_{\mathrm{gen}}\sim\mathcal N\!\left(\atanh\rho,\ \tfrac1N\right),\qquad
  z_{\mathrm{imp}}\sim\mathcal N\!\left(0,\ \tfrac1N\right).
  \label{eq:fisher}
\end{equation}
Because $\atanh$ is monotone, a threshold on $s$ and the corresponding threshold on $z$ produce identical error rates; the EER of the system is therefore the EER of the two normals in \eqref{eq:fisher}, which have equal variance and are separated by $\dprime=\sqrt N\,\atanh\rho$. Inserting into \eqref{eq:eer-dprime}:
\begin{equation}
  \boxed{\ \EER=\Phi\!\left(-\tfrac12\sqrt{N}\,\atanh\rho\right)\ }
  \label{eq:law}
\end{equation}
Fig.~\ref{fig:concept}b--c illustrates each step on simulated data. Nothing has been fitted.

\inwords{Compare two people through $N$ persistent features and you are, in effect, computing a correlation from $N$ paired numbers. The genuine correlation is centred on $\rho$, the impostor correlation on $0$, and both are uncertain by about $1/\sqrt N$ on Fisher's scale. The EER is the overlap of the two.}

Three qualitative properties of \eqref{eq:law} deserve notice before any numbers. First, the EER is $50\%$ at $\rho=0$ regardless of $N$ and tends to $0$ as $\rho\to1$ for any $N\ge1$. Second, the two quantities enter only through the product $\sqrt N\,\atanh\rho$: doubling $N$ is worth a factor $\sqrt2$ in $\dprime$, and $\atanh$ makes the last steps of persistence disproportionately valuable ($\atanh0.9=1.47$ but $\atanh0.99=2.65$). Third, the relation between EER and $\rho$ at fixed $N$ is not a straight line; its shape depends on $N$ (Fig.~\ref{fig:grid}), which is why the straight lines fitted in earlier work must be understood as local descriptions.

\subsection{What the Law Says About the Score Distributions}
\label{sec:scores}
The derivation also predicts the score distributions themselves, and these predictions can be compared with the observations of \cite{Friedman2020Sensors}. Transforming \eqref{eq:fisher} back to the cosine scale: the genuine median is $\rho$ and the impostor median is $0$; since the asymptotic variance of a sample correlation is $(1-\rho^2)^2/N$, the interquartile ranges are
\begin{equation}
  \mathrm{IQR}_{\mathrm{gen}}\approx1.349\,\frac{1-\rho^2}{\sqrt N},\qquad
  \mathrm{IQR}_{\mathrm{imp}}\approx\frac{1.349}{\sqrt N}.
  \label{eq:iqr}
\end{equation}
So raising persistence moves the genuine median up and narrows the genuine distribution, while the impostor distribution stays put---which is what \cite{Friedman2020Sensors} observed. Equation~\eqref{eq:iqr} makes the observation quantitative and, as Section~\ref{sec:simulation} shows, accurate.

\subsection{Exact Finite-$N$ Expression}
\label{sec:exact}
Fisher's normal approximation is the large-$N$ limit of an exact result, the density of the sample correlation coefficient \cite{Fisher1915,Hotelling1953} (Supplementary Material, Eq.~(S1)). Genuine scores follow that density with $\nu=N$ degrees of freedom and population correlation $\rho$, and impostor scores the same density with population correlation $0$. The exact EER is the crossing point of $\mathrm{FRR}(t)$ and $\mathrm{FAR}(t)$ computed from these densities, a one-dimensional root-finding problem (Supplementary Material, Section~S3). Section~\ref{sec:simulation} shows that it matches simulation at every $N$, including $N=2$.

For hand calculation, a finite-$N$ correction to \eqref{eq:law} is more convenient. Fisher's refined approximation gives a variance of $1/(\nu-2)$ rather than $1/\nu$, and $z$ is biased upward by about $\rho/(2\nu)$ \cite{Fisher1921}; the two effects pull the EER in opposite directions, and their net effect over the range studied is captured well by replacing $N$ with $N-1$ in \eqref{eq:law} (last column of Table~SI of the Supplementary Material). We refer to this as the finite-$N$ form and use it for all numerical predictions that follow unless stated otherwise.

\subsection{The Inverse Law and the Log-Linear Relation}
\label{sec:inverse}
Solving \eqref{eq:law} for $N$ gives the number of independent features required for a target EER,
\begin{equation}
  \boxed{\ N=\left(\frac{2\,q}{\atanh\rho}\right)^{2},\qquad q=\Phi^{-1}(1-\EER)\ }
  \label{eq:inverse}
\end{equation}
(add $1$ for the finite-$N$ form). Taking logarithms,
\begin{equation}
  \log_{10}N=2\log_{10}(2q)-2\log_{10}(\atanh\rho).
  \label{eq:loglinear}
\end{equation}
The EER target enters only through the first term, which is a constant offset; the second term is a fixed function of $\rho$. Over $0.35\le\rho\le0.95$ the function $-2\log_{10}(\atanh\rho)$ is close to a straight line of slope about $-2.2$ (between $-2.1$ and $-2.2$ with the finite-$N$ form, depending on the target). Equation~\eqref{eq:loglinear} is therefore a family of nearly parallel lines with slope $\approx-2$ and target-dependent intercepts---precisely the pattern reported empirically in \cite{Friedman2019Linear} (slopes $-1.99$ to $-2.09$).

\subsection{The Law in KCC Units}
\label{sec:kcc}
Combining \eqref{eq:kcc-spearman} with Pearson's relation gives the persistence $\rho$ that corresponds to a two-session KCC of $W$:
\begin{equation}
  \rho=2\sin\!\left(\frac{\pi(2W-1)}{6}\right)\approx2W-1,
  \label{eq:kcc2rho}
\end{equation}
and hence
\begin{equation}
  \boxed{\ \EER=\Phi\!\left(-\tfrac12\sqrt{N}\,\atanh\!\big(2W-1\big)\right)\ }
  \label{eq:law-kcc}
\end{equation}
up to the small correction in \eqref{eq:kcc2rho}. KCC values of 0.6, 0.7, 0.8, 0.9 and 0.95 correspond to $\rho$ of 0.21, 0.42, 0.62, 0.81 and 0.91.

Equations \eqref{eq:law} and \eqref{eq:law-kcc} rest on one simplification that the next section removes: every genuine comparison is assumed to share the same persistence $\rho$, so the genuine scores scatter only because $N$ is finite. Section~\ref{sec:subjects} lets persistence vary across comparisons, with spread $\tau$ on Fisher's scale; \eqref{eq:law} is the $\tau=0$ member of the resulting general form \eqref{eq:unified}.

\section{Extensions}
\label{sec:extensions}

\subsection{Features of Unequal Persistence}
\label{sec:unequal}
Real feature sets never share one ICC, and our earlier studies summarized a set by its \emph{median} ICC. The derivation suggests the \emph{mean}: the numerator of the genuine cosine score, $\sum_k x_{ik1}x_{ik2}/N$, has expectation $\bar\rho=\frac1N\sum_k\rho_k$, so the genuine score is centred on the mean persistence of the features, whatever their spread. Section~\ref{sec:sim-unequal} confirms that \eqref{eq:law} with $\rho=\bar\rho$ is accurate to about one EER point even for strongly bimodal persistence distributions, whereas the median can be wrong by an order of magnitude when the distribution is skewed. A wide spread of ICCs makes the law slightly optimistic---a second-order effect we do not model.

\subsection{Persistence That Varies Across Comparisons}
\label{sec:subjects}
Section~\ref{sec:unequal} lets persistence vary across features. The derivation of Section~\ref{sec:derivation} also assumes that every genuine comparison has the same population correlation $\rho$, so that the genuine scores scatter only because $N$ is finite. In practice each genuine comparison $c$ has its own persistence $\rho_c$. Two sources contribute. People differ: some are more consistent from one session to the next than others, the variation that speaker-recognition research has described as a ``menagerie'' of easy and difficult subjects \cite{Doddington1998}. And recordings differ: two captures of the same person may be well or poorly matched in pose, illumination, sensor placement, signal quality or elapsed time. The two cases behave differently from variation across features. When persistence varies across features, every genuine score is centred on the same mean persistence, and the law holds with the mean. When it varies across comparisons, the genuine Fisher-$z$ score of comparison $c$ is $\mathcal N(\atanh\rho_c,1/N)$, so the genuine distribution is a mixture of normals, while the impostor distribution $\mathcal N(0,1/N)$ is unchanged. The EER is the threshold $t$ at which
\begin{equation}
  \underbrace{\E_c\,\Phi\!\big(\sqrt N\,(t-\atanh\rho_c)\big)}_{\mathrm{FRR}(t)}
  =\underbrace{\Phi\!\big(-\sqrt N\,t\big)}_{\mathrm{FAR}(t)},
  \label{eq:mixture}
\end{equation}
a one-dimensional root that reduces to \eqref{eq:law} when all $\rho_c$ are equal. Comparisons of low persistence populate the lower tail of the genuine distribution, and it is that tail which sets the EER, so the law evaluated at the mean persistence is optimistic, increasingly so as the spread grows (Section~\ref{sec:sim-subjects}).

When the spread is moderate and unimodal, a closed form follows. Let $\atanh\rho_c$ have mean $m$ and standard deviation $\tau$ across comparisons and be approximately normal. The genuine Fisher-$z$ scores are then $\mathcal N(m,\tau^2+1/N)$, and the EER is
\begin{equation}
  \boxed{\ \EER=\Phi\!\left(-\,\frac{m}{\sqrt{\tau^2+1/N}+1/\sqrt N}\right)\ }
  \label{eq:unified}
\end{equation}
With $\tau=0$ it is \eqref{eq:law}, with $m=\atanh\rho$. As $N$ grows the denominator falls to $\tau$, so
\begin{equation}
  \EER\;\longrightarrow\;\Phi(-m/\tau)\qquad(N\to\infty),
  \label{eq:floor}
\end{equation}
a floor that no number of features can lower. Equation~\eqref{eq:unified} therefore has two limiting regimes, set by the ratio of the persistence spread to the sampling spread, $\tau\sqrt N$. When $\tau\sqrt N\ll1$ the error is limited by the number of independent features, and adding or decorrelating features lowers it as Section~\ref{sec:derivation} describes. When $\tau\sqrt N$ is about $3$ or more, most of the attainable separation has been reached and the error is limited by the spread of persistence: only raising persistence, or making it more uniform across comparisons, helps much. Between the two, both terms matter. For correlated features $N$ is replaced by $\Neff$ throughout (Section~\ref{sec:neff}); in applications the sampling variance $1/N$ becomes $v=1/(\Neff-2)$, the variance of the impostor Fisher-$z$ scores, which \eqref{eq:neff-imp} measures directly.

The spread can be estimated from two sessions. By the law of total variance $\Var(\atanh s_{\mathrm{gen}})\approx1/N+\tau^2$, while the impostor variance is $1/N$, so the excess of the genuine over the impostor variance estimates $\tau^2$. The same excess arises when session noise is shared across features (Section~\ref{sec:sim-corr}), so it is an upper bound unless that cause can be excluded. Written in terms of the score distributions, \eqref{eq:unified} is the two-normal form of Section~\ref{sec:session} with $\mu_i=0$, $\sigma_i=1/\sqrt N$ and $\sigma_g=\sqrt{\tau^2+1/N}$: the two-normal form is the Gaussian case of this extension, and the equal-spread law is its case $\tau=0$. Two moments do not describe a heavy lower tail, however, and when a minority of comparisons has very low persistence \eqref{eq:mixture} must be evaluated with the distribution of $\rho_c$ itself. If the comparisons of one person share a persistence $\rho_i$ and three or more sessions are available, $\rho_i$ can be estimated subject by subject from session pairs other than the one tested. If the spread lies mainly between recordings, the genuine Fisher-$z$ scores of pilot subjects, shrunk toward their mean by the factor $\sqrt{\tau^2/\Var(\atanh s_{\mathrm{gen}})}$ to remove the sampling part, estimate the distribution of $\atanh\rho_c$. Variation in persistence also biases the inverse law: an effective number of features inferred from an observed EER at the mean persistence is too small, because part of the observed error is due to the spread of persistence rather than to a lack of features.

\inwords{If some comparisons are much less consistent than others---some people, or some pairs of recordings---the error rate is set by them, not by the average comparison. Averaging the persistence first and then applying the law underestimates the error. Once there are enough features, the spread of persistence, not the number of features, sets the error.}

\subsection{Intercorrelated Features: the Effective Number}
\label{sec:neff}
Suppose the $N$ standardized features are correlated within a session, with correlation matrix $\mathbf R$ (we write $r_{jk}$ for the correlation between two features and reserve $\rho$ for the test--retest correlation of one feature). The impostor cosine score is, to leading order, $\mathbf x_1^{\!\top}\mathbf x_2/N$ for two independent vectors with covariance $\mathbf R$, and
\begin{equation}
  \Var\!\big(\mathbf x_1^{\!\top}\mathbf x_2\big)=\tr(\mathbf R^2)=N+\sum_{j\ne k}r_{jk}^2
\end{equation}
(Supplementary Material, Section~S2). Hence $\Var(s_{\mathrm{imp}})\approx\tr(\mathbf R^2)/N^2$, which is $1/N$ for independent features and larger otherwise. Defining
\begin{equation}
  \boxed{\ \Neff=\frac{N^2}{\tr(\mathbf R^2)}=\frac{N}{1+(N-1)\,\overline{r^2}}\ }
  \label{eq:neff}
\end{equation}
where $\overline{r^2}$ is the mean squared off-diagonal correlation, the impostor spread is that of $\Neff$ independent features. $\Neff$ is the participation ratio of the eigenvalues of $\mathbf R$, $(\sum_k\mu_k)^2/\sum_k\mu_k^2$, a standard measure of the effective number of independent variables \cite{Bretherton1999} (related definitions based on the eigenvalue spread are used for correlated statistical tests \cite{Cheverud2001,Nyholt2004}). Two properties matter for design. Equation~\eqref{eq:neff} needs only the mean squared pairwise correlation, which is routinely measured for any feature set or embedding; if a study reports the mean and standard deviation of the absolute correlations, $\overline{r^2}=(\overline{|r|})^2+\mathrm{SD}^2$. And as $N$ grows, $\Neff\to1/\overline{r^2}$: with a mean absolute correlation of $0.19$ and a spread of $0.14$, as reported for the embeddings of \cite{Raju2026TBIOM} (Spearman correlations, used here as a proxy for Pearson), $\overline{r^2}\approx0.19^2+0.14^2=0.056$, and no more than about $18$ effective features can be obtained however many are extracted.

Substituting $\Neff$ for $N$ in \eqref{eq:law} is a first-order rule. It is not exact because the normalization by the vector norms in \eqref{eq:cosine} interacts with the correlation structure, and because the genuine variance depends on where the correlation lives (in the trait part, in the noise part, or in both). Section~\ref{sec:sim-corr} quantifies its accuracy: within $3.5$ EER points for mean pairwise correlations up to $0.2$, and pessimistic by several points when one common factor dominates the correlation matrix.

A calibration-free alternative that needs no correlation matrix follows from \eqref{eq:fisher}: the impostor Fisher-$z$ scores of $\Neff$ independent features have variance $1/(\Neff-2)$ by Fisher's refined approximation, so
\begin{equation}
  \widehat N_{\mathrm{eff}}=2+\frac{1}{\Var\!\big(\atanh s_{\mathrm{imp}}\big)}.
  \label{eq:neff-imp}
\end{equation}
This is Daugman's degrees-of-freedom argument \cite{Daugman1993} transplanted from Hamming distances to cosine scores. In simulation it agrees with \eqref{eq:neff} to within about $15\%$ on average, and where the two disagree it is the more accurate (Section~\ref{sec:sim-corr}).

\inwords{Correlated features share their information. Count how much the impostor scores spread: the spread is that of $\Neff$ independent features, and $\Neff$ can never exceed one over the mean squared correlation.}

\subsection{Decorrelation and the Canonical Persistence Spectrum}
\label{sec:whitening}
Zero-phase (ZCA) whitening \cite{Kessy2018} turns the $N$ correlated features into uncorrelated unit-variance components. Rotating these to the eigenvectors of their cross-session covariance matrix gives $N$ components whose persistences $\lambda_1,\dots,\lambda_N$, the \emph{canonical persistences}, are invariant to any invertible linear transformation of the features (derivation in Section~S4 of the Supplementary Material). Cosine scoring of the whitened features is then cosine scoring of $N$ independent features, so $\EER_{\mathrm{whitened}}\approx\Phi(-\tfrac12\sqrt N\,\atanh\bar\lambda)$, with $\bar\lambda=\tr(\boldsymbol\Sigma^{-1}\mathbf C)/N$ and $\boldsymbol\Sigma$ and $\mathbf C$ the within-session and cross-session covariance matrices. In simulation this is accurate to well under one EER point (Section~\ref{sec:sim-corr}). A system that whitens its features is described by this form with the full $N$, whereas \eqref{eq:neff} describes what raw cosine scoring delivers.

\subsection{Session-Level Changes and the Choice of Index}
\label{sec:session}
Model \eqref{eq:model} assumes that session~2 differs from session~1 only by fresh noise. A real second session---or a signal manipulation applied to one session---may also add a per-dimension shift, a uniform rescaling of the whole vector, or a per-dimension rescaling. Cosine similarity ignores a uniform rescaling but is hurt by shifts and by dimension-specific rescaling. Pearson $r$ and KCC are blind to all three, because they are invariant to linear (KCC: monotone) changes of a dimension; $\ICC(2,1)$ is reduced by all three, because it measures absolute agreement. No single index is therefore right for every kind of session change (Section~\ref{sec:sim-session}). The unambiguous procedure is to read the two quantities the law needs directly off the score distributions: $\hat\rho$ = median genuine score, which by construction is the persistence of whatever enters the cosine, and $\widehat N_{\mathrm{eff}}$ from \eqref{eq:neff-imp}. When the genuine and impostor spreads differ---which happens when session noise is shared across features, or when persistence varies across comparisons (Section~\ref{sec:subjects})---the general two-normal form \cite{Poh2005} $\EER=\Phi\big(-(\mu_g-\mu_i)/(\sigma_g+\sigma_i)\big)$, with means and standard deviations taken in Fisher-$z$ units, replaces \eqref{eq:eer-dprime}; it is still free of fitted parameters. The persistence indices then have a separate, well-posed job---predicting $\hat\rho$ from single-dimension statistics---and the simulations say which will succeed when.

\subsection{Averaged Enrollment}
\label{sec:averaging}
Deep eye-movement pipelines average the embeddings of $k$ short sequences to form both the enrollment centroid and the probe centroid ($k=12$ in \cite{Raju2026TBIOM}). If the sequence noises were independent, the persistence of the centroids would be the Spearman--Brown value $\rho_k=k\rho/[1+(k-1)\rho]$ \cite{Spearman1910,Brown1910}. Sequences within a session share session-level noise, so in practice the gain saturates, which is the diminishing return beyond about five sequences reported in \cite{Raju2026TBIOM}. The law must be fed the persistence of the vectors that enter the cosine, i.e.\ of the centroids (Supplementary Material, Section~S5).

\section{Simulation Study}
\label{sec:simulation}

\subsection{Methods}
All simulations follow the protocol of our earlier synthetic-feature work. Features are generated from \eqref{eq:model}, optionally with correlation matrices imposed on the trait and noise parts, and are $z$-scored across the pooled sessions. Every subject's session-1 vector is compared with every subject's session-2 vector by cosine similarity; the $n$ diagonal comparisons are genuine and the $n(n-1)$ off-diagonal ones impostor. The EER is read off the crossing of $\mathrm{FRR}(t)=P(s_{\mathrm{gen}}<t)$ and $\mathrm{FAR}(t)=P(s_{\mathrm{imp}}\ge t)$ with linear interpolation; impostor scores are accumulated in a histogram with bin width $10^{-4}$ so that $n=10{,}000$ subjects ($10^8$ impostor scores) can be handled. Random seeds are fixed; all scripts run under MATLAB or GNU Octave without toolboxes.

\subsection{The $N\times\ICC$ Surface}
\begin{figure*}[t]
\centering
\includegraphics[width=\textwidth]{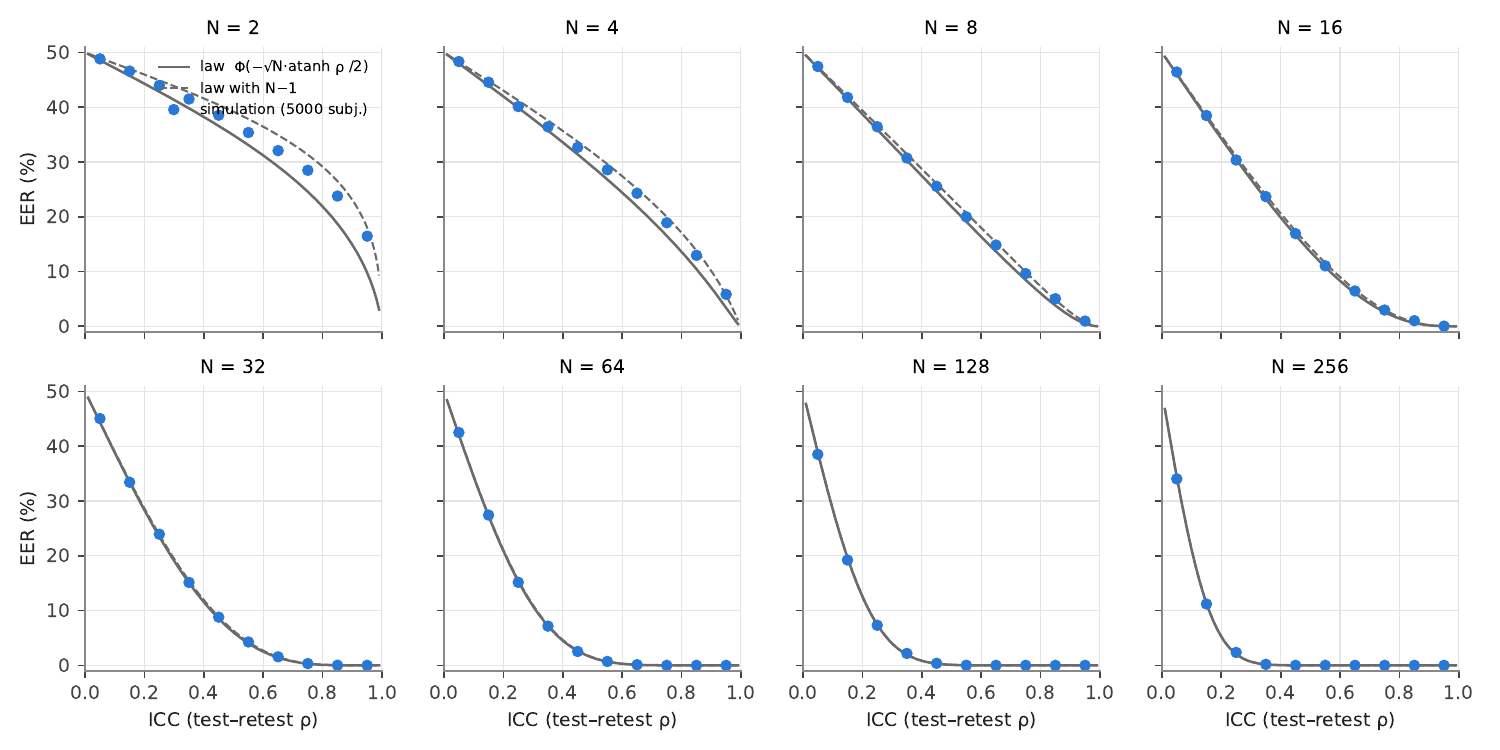}
\caption{Simulated EER (5,000 subjects, dots) versus the law \eqref{eq:law} (solid) and its finite-$N$ form with $N-1$ (dashed), for $N=2$ to $256$ independent features and $\ICC=0.05$ to $0.95$. The panels also show why no single straight line can describe the EER--ICC relation: its shape depends on $N$.}
\label{fig:grid}
\end{figure*}
Fig.~\ref{fig:grid} compares \eqref{eq:law} with simulation over $N\in\{2,4,\dots,256\}$ and $\ICC\in\{0.05,\dots,0.95\}$ (5,000 subjects per cell). For $N\ge16$ the closed form is within $0.7$ EER points everywhere; for $N\le8$ it is optimistic at high persistence (e.g.\ $N=8$, $\ICC=0.85$: $3.8\%$ predicted, $5.0\%$ simulated), and the finite-$N$ form removes most of the discrepancy. The score-distribution predictions of Section~\ref{sec:scores} are met precisely: at $N=32$ the genuine medians are $0.047,0.149,\dots,0.952$ for $\ICC=0.05,0.15,\dots,0.95$; the impostor IQR is $0.243$ at every ICC against $0.238$ predicted by \eqref{eq:iqr}; and the genuine IQR falls from $0.238$ to $0.024$ against $0.238$ to $0.023$ predicted.

\subsection{Exact Versus Approximate}
Table~SI of the Supplementary Material shows that the exact expression reproduces the simulation at every $N$ down to $N=2$, where the normal approximation fails badly ($9.8\%$ against $16.3\%$ at $\ICC=0.95$). The finite-$N$ form is a good compromise from $N=4$ upward.

\subsection{Number of Features for a Target EER}
\label{sec:sim-target}
In \cite{Friedman2019Linear} the number of independent features needed to bring the EER below $5,2,1,0.5$ and $0.1\%$ was searched for by simulation (10,000 subjects, 100 repetitions) in seven ICC bands. Table~SII of the Supplementary Material (Section~S7) lists those counts, the simulated EER we obtain at exactly those counts, and the prediction of the inverse law \eqref{eq:inverse} in its finite-$N$ form. The re-simulated EER is at or just below the target in every cell (overshoots of at most $0.14$ points; the undershoots of up to $1.3$ points in the $0.95$ band reflect the coarse steps that integer feature counts impose when $N$ is small), confirming that the two pipelines agree. The inverse law predicts the required count to within two features in 34 of the 35 cells; the exception is the $0.1\%$ target at $\ICC=0.35$, where the law is high by $6$ features out of $281$. Fig.~S1 of the Supplementary Material shows the same comparison on the logarithmic scale of \cite{Friedman2019Linear}: the five near-parallel lines of that paper are the curves of \eqref{eq:loglinear}.

\subsection{Unequal Persistence}
\label{sec:sim-unequal}
Table~SIII of the Supplementary Material tests Section~\ref{sec:unequal}. For $N=32$ features whose ICCs are spread uniformly or bimodally, the law evaluated at the mean ICC is within $1.1$ EER points of the simulation in every case; evaluated at the median it fails whenever the distribution is skewed (e.g.\ $19.4\%$ predicted against $9.1\%$ observed when three quarters of the features have $\ICC=0.3$ and one quarter $\ICC=0.9$). The residual with the mean is always in the same direction---the law is slightly optimistic when the spread is large.

\subsection{Persistence Varying Across Subjects}
\label{sec:sim-subjects}
Table~SIV of the Supplementary Material tests Section~\ref{sec:subjects}. Each of 4,000 simulated subjects has its own persistence $\rho_i$, shared by all $N=32$ independent features, drawn either from a normal distribution with mean $0.6$ or as a two-point mixture in which a minority of subjects has $\rho=0.2$ and the rest $\rho=0.7$. The law at the mean persistence stays near $2.7\%$ whatever the spread, while the simulated EER rises to $10\%$: at a standard deviation of $0.15$ across subjects the law is optimistic by a factor of $2.2$, and with $10\%$ low-persistence subjects by a factor of $4.3$. The two-normal form errs in both directions, too high for the normal spreads and too low for the two-point mixtures. The mixture form \eqref{eq:mixture}, evaluated at the simulated $\rho_i$, is within $0.5$ EER points of the simulation in every row. The last column shows the effective number of features that the inverse law infers from the simulated EER at the mean persistence: $32$ independent features are read as $21$ at a spread of $0.15$ and as $14$ when a fifth of the subjects have low persistence.

\subsection{Non-Normal Features and the Range of the KCC Route}
\label{sec:sim-kcc}
Kendall's $W$ is invariant to any monotone change of a feature, and cosine similarity is not, so the KCC route \eqref{eq:law-kcc} is blind to exactly the departure from normality that changes the EER (details in Section~S10 of the Supplementary Material). For $N=32$ features with $\rho=0.6$, made heavier-tailed without changing any rank, the law in KCC units is exact at excess kurtosis $0$ and optimistic by $0.6$ EER points at $2.9$ and by $6.9$ at $43$. Two practical rules follow: read the law in KCC units only while the typical dimension has excess kurtosis of about $1$ or less, and, when it does not, replace each feature by the inverse-normal scores of its ranks before scoring---a monotone change that leaves the KCC unaltered and restored the law to within $0.02$ EER points in the heaviest-tailed case tested.

\subsection{Intercorrelated Features}
\label{sec:sim-corr}
We imposed an exchangeable correlation on $N=32$ features with $\ICC=0.5$ or $0.7$ in four ways: the same correlation in trait and noise (A), only among the traits (B), only among the noise (C), and four independent blocks (D); Section~S11 of the Supplementary Material gives the details and Fig.~S2. In every case the per-feature ICC is unchanged, yet the EER of raw cosine scoring rises steeply with the correlation. The rule with $\Neff$ from \eqref{eq:neff} is within $1.3$ points when the mean within-session $|r|$ is below $0.1$, within $3.5$ points up to $0.2$, and pessimistic by $4$--$11$ points when a single common factor dominates. After ZCA whitening, the law with the mean canonical persistence is within $0.4$ points in all 48 conditions. The two-normal form with the impostor-based estimate \eqref{eq:neff-imp} and separate spreads is within $1.2$ points in all 32 correlated conditions tested; the ratio $\sigma_g/\sigma_i$ is one for independent features and exceeds one when session noise is shared across features.

\subsection{Session-Level Changes}
\label{sec:sim-session}
Table~SV of the Supplementary Material (Section~S12) applies the session changes of Section~\ref{sec:session} to session~2 of an $N=32$, $\ICC=0.7$ set. A shift of one standard deviation in every dimension raises the EER from $0.8\%$ to $6.4\%$, and only $\ICC(2,1)$ registers it ($0.70\to0.47$). A uniform rescaling by $3$ leaves the EER at $0.8\%$ but drives $\ICC(2,1)$ to $0.42$, so the law evaluated with it is far too pessimistic ($10.6\%$), while Pearson $r$ and KCC correctly report no change. Evaluated with $\hat\rho$ and $\widehat N_{\mathrm{eff}}$ read off the score distributions, the law and the two-normal form are within $0.4$ points in every row, whatever the session change.

\section{The Law on Real Data}
\label{sec:realdata}
Two published results from our own programme allow the law to be confronted with real biometric performance: one summarized by median ICCs, the other by condition-level KCC and EER. Neither summary is a sharp test, but both are informative and both were produced before the law was derived; for the second, the stored embeddings allow a further test at the level of the score distributions themselves (Sections~\ref{sec:ekyt-scores}--\ref{sec:ekyt-saturation}). Three further modalities, unrelated to ours, test whether anything in the law is specific to eye movements: keystroke dynamics, the ECG, and faces (Sections~\ref{sec:keystroke}--\ref{sec:face}). They also span the range of the general form of Section~\ref{sec:subjects}: in keystroke dynamics the spread of persistence is small beside the sampling spread, in the ECG it varies between people, and in face photographs between recordings.

\subsection{Twenty-Nine Feature Subsets, Twelve Datasets}
\label{sec:plos}
\begin{figure*}[t]
\centering
\includegraphics[width=\textwidth]{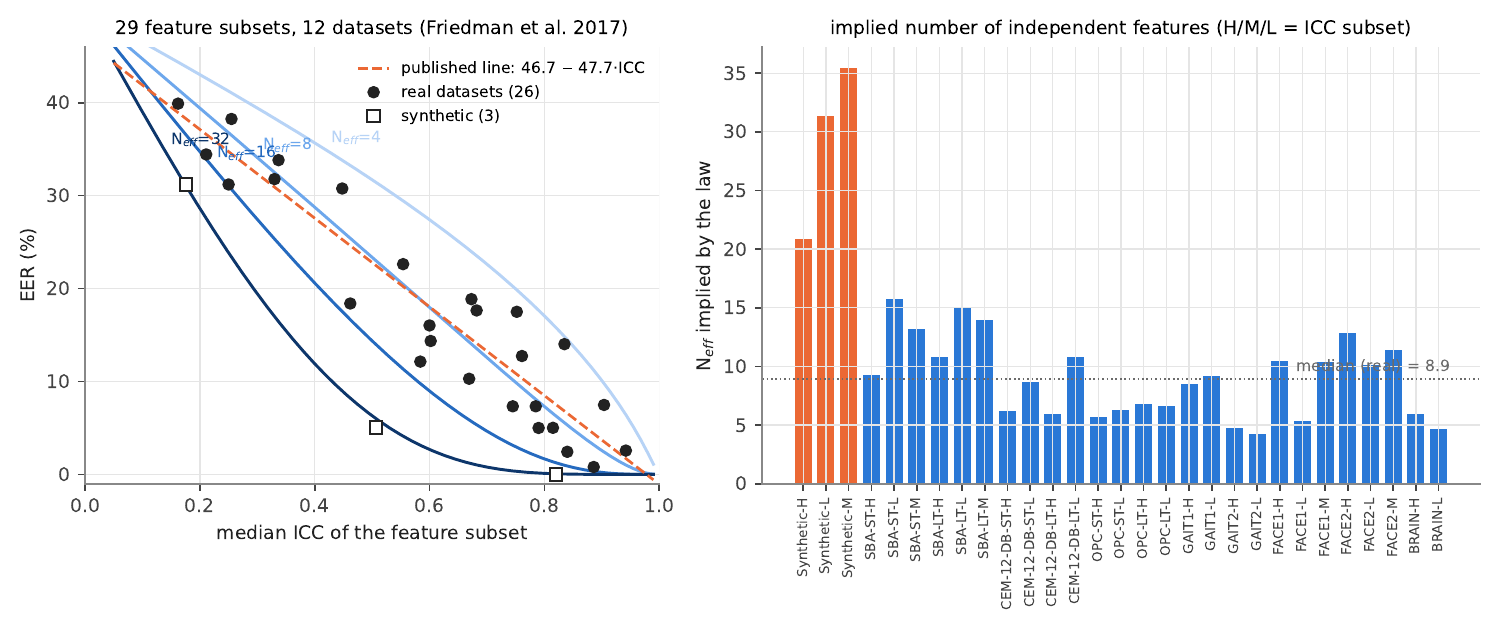}
\caption{The law on the 29 feature subsets of \cite{Friedman2017PLoS}. Left: EER versus median ICC of each subset, the published straight-line fit, and the law's curves for $\Neff=4,8,16,32$ (finite-$N$ form). Right: the effective number of independent features implied by the law for each subset (H/M/L: high-, moderate-, low-ICC subset; orange: the synthetic set).}
\label{fig:plos}
\end{figure*}

In \cite{Friedman2017PLoS}, hand-crafted feature sets from six eye-movement, two gait, two face and one brain-structure dataset, plus one synthetic set, were split into low-, moderate- and high-ICC subsets (29 subsets in all, 26 from real data and 3 synthetic), reduced by PCA on training subjects, and scored by cosine similarity on held-out subjects. For each subset the median ICC of its features and the EER are known. Inverting the law gives the number of independent features that would produce that EER at that persistence,
\begin{equation}
  \Neff=\left(\frac{2\,q}{\atanh(\text{median ICC})}\right)^2+1,\qquad q=\Phi^{-1}(1-\EER),
\end{equation}
which is a property of the pipeline rather than of the feature count. Fig.~\ref{fig:plos} shows the result. For the 26 real-data subsets $\Neff$ is small and stable: median $8.9$, interquartile range $6.0$--$10.8$, range $4$--$16$, although the subsets contained $8$--$122$ features and $3$--$22$ PCA components were retained. The three synthetic subsets, whose features are nearly independent, give $\Neff=21$ (a lower bound, since an EER of $0$ was floored at $0.5\%$) to $35$, in line with the 30 components extracted from them. Real feature sets thus behave, under PCA and cosine scoring, like five to fifteen independent features: their variance is concentrated in a few retained components, and cosine similarity weights components by variance, so the participation ratio of the component variances---and with it $\Neff$---is small. This reading attributes the whole shortfall to effective dimension. Part of it may instead come from persistence that varies across subjects or recordings, which the inverse law also registers as a smaller $\Neff$ (Section~\ref{sec:subjects}): in Table~SIV of the Supplementary Material, $32$ independent features are read as $21$ when per-subject persistence has a standard deviation of $0.15$, and as $14$ when a fifth of the subjects have low persistence. The values in Fig.~\ref{fig:plos} therefore tend to understate the effective dimension of these pipelines. The published summaries cannot separate the two effects; Stage~5 of Section~\ref{sec:limitations} can.

This also explains the ``biometric law'' $\EER=50-50\,\ICC$. Over the range of persistence that real subsets occupy ($0.15$--$0.95$), the curve \eqref{eq:law} for $\Neff\approx8$ is nearly straight, running from about $42\%$ to about $1\%$; a least-squares line through it has slope $-53$ and intercept $50$. The straight line is therefore a property of the typical effective dimensionality of PCA-reduced feature sets, not a universal relation; a pipeline with more effective features shows a steeper, convex relation that reaches $\EER\approx0$ well before $\ICC=1$ (Fig.~\ref{fig:grid}, $N\ge32$).

\subsection{Deep Eye-Movement Embeddings}
\label{sec:ekyt}
In \cite{Raju2026TBIOM}, 128-dimensional embeddings of the Eye Know You Too network \cite{Lohr2022EKYT}, which has also been benchmarked at scale for gaze-based authentication in virtual reality \cite{Lohr2024VR}, were computed for the GazeBase \cite{Griffith2021GazeBase} and GazeBaseVR \cite{Lohr2023GazeBaseVR} datasets under systematic degradations of the eye-movement signal (two forms of down-sampling, additive noise, and shorter recordings), and for each condition the KCC of the embedding dimensions, the EER of cosine-similarity authentication, and the mean absolute Spearman intercorrelation among the dimensions were reported. The intercorrelation stayed at $0.19$ (SD $0.14$) under all manipulations. From \eqref{eq:neff}, $\overline{r^2}\approx0.19^2+0.14^2=0.056$ and
\begin{equation}
  \Neff=\frac{128}{1+127\times0.056}\approx16 .
\end{equation}
The law therefore makes a prediction with no free parameter: the KCC--EER points should fall on \eqref{eq:law-kcc} with $\Neff-1=15$.

\begin{figure*}[t]
\centering
\includegraphics[width=0.9\textwidth]{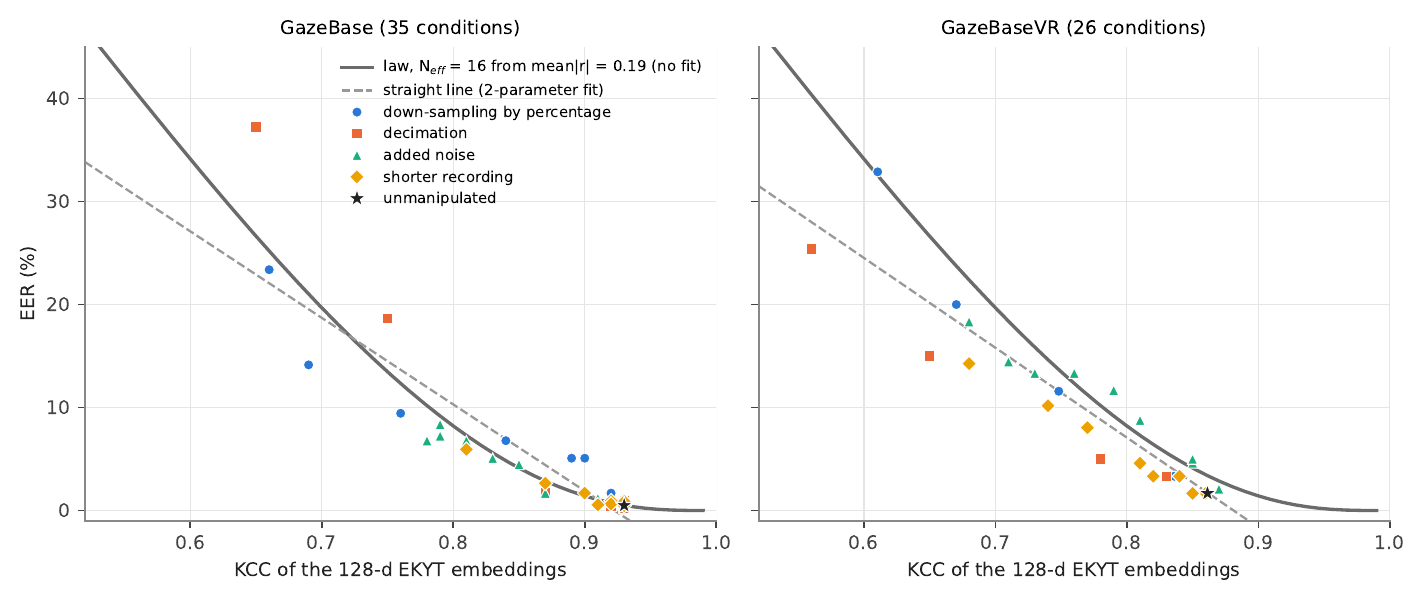}
\caption{The 61 conditions of \cite{Raju2026TBIOM} (reading task; published per-condition KCC and EER). Left: 35 GazeBase conditions. Right: 26 GazeBaseVR conditions. Solid: the law \eqref{eq:law-kcc} at $\Neff=16$, computed a priori from the published intercorrelation (finite-$N$ form); dashed: a two-parameter straight line fitted to each dataset. Markers show the manipulation; the star is the unmanipulated recording.}
\label{fig:ekyt}
\end{figure*}

We use the per-condition KCC and EER published with that study for the reading (TEX) task, on which its 61 conditions are defined: 35 GazeBase conditions (seven levels each of down-sampling by percentage and of decimation, nine noise levels, eleven shorter recordings, and the unmanipulated recording) and 26 GazeBaseVR conditions (four, four, nine and eight levels, and the unmanipulated recording). Fig.~\ref{fig:ekyt} shows the comparison. On GazeBase the law with $\Neff=16$ explains $R^2=0.88$ of the EER variance, and the least-squares $\Neff$ is $16.0$: the value computed a priori from the intercorrelation is also the value that fits best. A two-parameter straight line fitted to the same points explains $R^2=0.80$. The $\Neff$ implied point by point has median $16.1$ (interquartile range $13.6$--$18.1$; the one EER below $0.3\%$ was floored at $0.3\%$ before inversion, since an EER of $0$ cannot be inverted). A one-parameter curve predicted from an independently measured quantity thus fits better than the fitted line, and it follows the curvature at both ends of the range that a line cannot. On GazeBaseVR the law is pessimistic: most conditions fall below the curve, most clearly under decimation (on average $8.5$ EER points below) and for shorter recordings ($3.4$ points below). There the least-squares $\Neff$ is $21.6$, and the law explains $R^2=0.63$ against $0.87$ for the straight line. The departure is in the direction of more effective features, or more persistence entering the cosine, than the intercorrelation and the KCC indicate; Section~\ref{sec:ekyt-index} separates the two and finds that it is $\Neff$ that moves.

\subsection{Score Distributions of the Embeddings}
\label{sec:ekyt-scores}
The published summaries give one KCC and one EER per condition. The embeddings of that study were stored, so the score distributions that the derivation is about can be examined directly; this is Stage~1 of the protocol in Section~\ref{sec:limitations}. We took the stored enrollment and probe templates of that study---the centroids of 12 (GazeBase) or 9 (GazeBaseVR) reading-task sequences of the first round---and scored them by cosine similarity, every enrollment template against every probe. Of the 61 conditions we use the 56 (30 GazeBase, 26 GazeBaseVR) whose EER, recomputed from the stored templates, agrees with the published value to within one genuine error---$1/59$ on GazeBase and $1/60$ on GazeBaseVR, about $1.7$ EER points---which is the resolution the subject count allows; the median agreement is $0.1$ points. Five GazeBase conditions---decimation to 25 and 10\,Hz and the three highest noise levels---are excluded because no archived version of the embeddings reproduces their published EER. The mean KCC of the stored templates agrees with the published value to within $0.02$ in 54 of the 56, the exceptions being GazeBaseVR decimation to 25 and 10\,Hz ($0.06$ and $0.10$); the analyses below use the recomputed indices throughout, so they are internally consistent.

The first finding is that the stored embeddings are not centred. They share a large common component, so the impostor scores have a median Fisher-$z$ mean of $0.63$ across conditions rather than $0$. This is the case that Section~\ref{sec:limitations} warns about, and the equal-spread law read off these scores fails outright ($R^2<0$). We therefore centred each condition on the mean of its own templates, because the embeddings of the training subjects are not available. The impostor scores are then centred (median Fisher-$z$ mean $-0.01$), and the law with $\hat\rho$ the median genuine score and $\widehat N_{\mathrm{eff}}$ from the impostor spread \eqref{eq:neff-imp} explains $R^2=0.94$ of the EER measured on the same scores. The two-normal form of Section~\ref{sec:session} explains $R^2=0.98$, with a median error of $0.63$ EER points (Fig.~S3a of the Supplementary Material); on the $\dprime$ scale its predictions have slope $0.84$ against the measured values.

The second finding concerns $\Neff$. Read off the impostor spread of the centred scores, it has median $16.8$ (interquartile range $14.7$--$17.6$), close to the value predicted a priori from the published intercorrelation. Computed from the measured intercorrelation of the 128 dimensions by \eqref{eq:neff}, it has median $14.0$ (Fig.~S3b of the Supplementary Material); the impostor-based value is the larger, as in the simulations of Section~\ref{sec:sim-corr}. The genuine Fisher-$z$ spread exceeds the impostor spread by a median factor of $1.30$, which is why the two-normal form does better than the equal-spread law. By Section~\ref{sec:subjects} this excess bounds the spread of persistence across comparisons, $\tau$, at about $0.21$ on Fisher's scale (median over conditions), an upper bound because shared session noise produces the same excess.

Across the 56 conditions the two-normal prediction explains $R^2=0.94$ of the \emph{published} EERs as well, so the agreement does not depend on using our own EERs as the target.

\subsection{Which Index Predicts the Score Centre, and What Moved on GazeBaseVR}
\label{sec:ekyt-index}
Stage~2 of the protocol asks which per-dimension persistence index predicts $\hat\rho$, the centre of the genuine score distribution (details and Table~SVI in Section~S14 of the Supplementary Material). Over the 56 conditions the mean Pearson $r$, $\ICC(2,1)$ and KCC track $\hat\rho$ equally well (correlations $0.992$--$0.993$) and understate it by $0.030$--$0.043$, because the cosine weights each dimension by its variance; weighting the Pearson $r$ by variance reduces the bias to $-0.013$. The index is not why the GazeBaseVR conditions fall below the curve in Fig.~\ref{fig:ekyt}: the KCC offset there is the same as everywhere else. It is $\Neff$, which decimation and shortened recordings raise above the fixed value of $16$ that the curve assumes, and the score-based form of Section~\ref{sec:ekyt-scores} predicts these conditions within one EER point.

\subsection{Does the Effective Number Saturate?}
\label{sec:ekyt-saturation}
The design claim of this paper---that intercorrelation caps what extra dimensions can buy---has so far been an extrapolation of \eqref{eq:neff}. Stage~3 tests it. From the centred templates of the unmanipulated recording and of the mid-level noise condition of each dataset, we drew random subsets of $4,8,16,32,64,96$ and $128$ of the $128$ embedding dimensions (25 draws per size), and scored each subset exactly as before.

\begin{figure*}[t]
\centering
\includegraphics[width=0.9\textwidth]{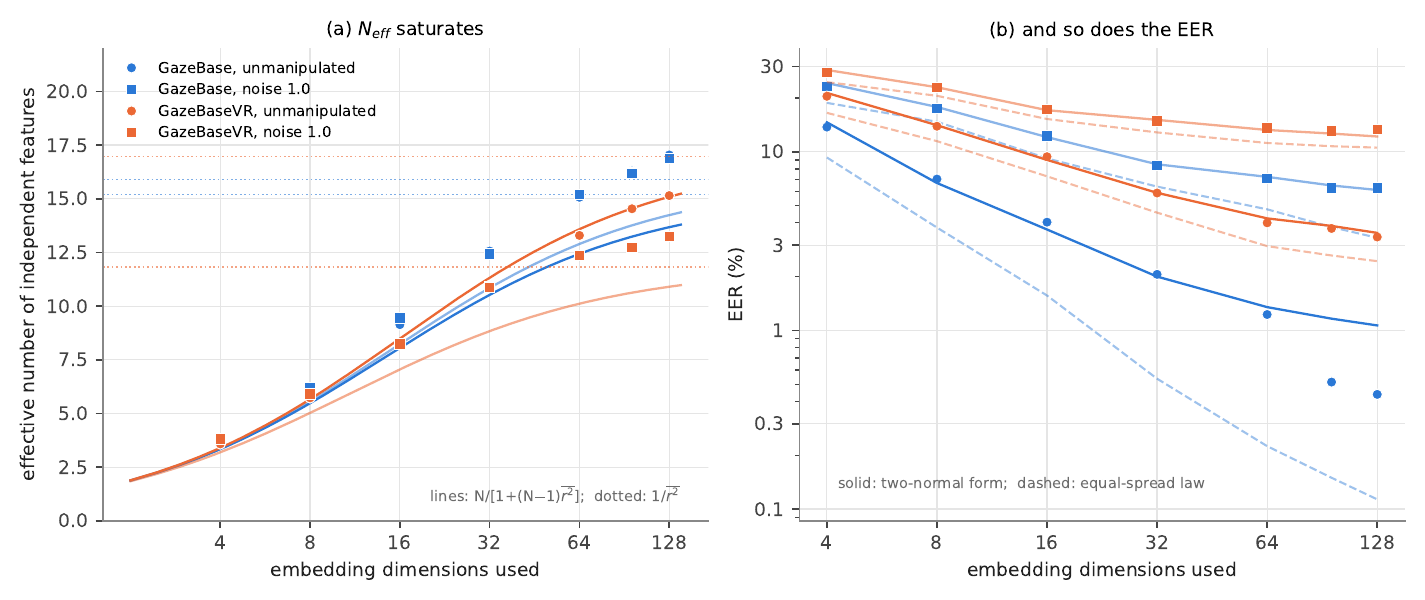}
\caption{Random subsets of the 128 embedding dimensions, 25 draws per size, centred templates. (a) Effective number of independent features from the impostor spread \eqref{eq:neff-imp} (markers, mean over draws) against the number of dimensions used; lines: $N/[1+(N-1)\overline{r^2}]$ with the measured $\overline{r^2}$ of each condition; dotted: the ceiling $1/\overline{r^2}$. (b) Measured EER (markers) with the two-normal form (solid) and the equal-spread law with $\hat\rho$ and $\widehat N_{\mathrm{eff}}$ (dashed). Both axes logarithmic.}
\label{fig:saturation}
\end{figure*}

Fig.~\ref{fig:saturation}a shows the result. The effective number of independent features grows far more slowly than the number of dimensions: averaged over draws, 8 dimensions behave like $5.7$--$6.2$ independent features, 16 like $8.2$--$9.5$, 32 like $10.9$--$12.6$, 64 like $12.4$--$15.2$, and all 128 like $13.2$--$17.0$. Doubling from 64 to 128 dimensions adds between $7\%$ and $14\%$. The growth follows $N/[1+(N-1)\overline{r^2}]$ with the measured $\overline{r^2}$ of each condition, and flattens towards $1/\overline{r^2}$, which is $11.8$--$17.0$ here. Two independent estimates agree on this: the impostor spread \eqref{eq:neff-imp} and the intercorrelation \eqref{eq:neff} differ by less than $3$ effective features at every size. The saturation predicted in Section~\ref{sec:design} is therefore a property of these embeddings and not an artefact of the first-order rule.

Fig.~\ref{fig:saturation}b shows what that costs in error rate, and adds a second effect. The EER falls steeply up to about 32 dimensions and then flattens: on GazeBaseVR with noise, from $14.8\%$ at 32 dimensions to $13.3\%$ at 128; on unmanipulated GazeBase, from $2.1\%$ to $0.44\%$. The equal-spread law with $\hat\rho$ and $\widehat N_{\mathrm{eff}}$ (dashed) tracks the measurements while the dimensions are few but becomes optimistic as they grow---by a factor of four at 128 dimensions on unmanipulated GazeBase. The two-normal form (solid) stays with the measurements throughout, with a median error of $0.70$ EER points over all 604 subsets and $0.40$ at 128 dimensions, against $2.10$ and $1.84$ for the equal-spread law. The reason is visible in the score distributions: the ratio of the genuine to the impostor Fisher-$z$ spread rises from $1.02$ at 4 dimensions to $1.49$ at 128 on unmanipulated GazeBase. As dimensions are added, the $1/N$ measurement noise shrinks while the spread of persistence across comparisons does not, so the $\tau$ term of \eqref{eq:unified} comes to dominate the genuine distribution and sets a floor that no further dimension can lower. Adding dimensions to a learned embedding therefore runs into two limits at once, and a design that ignores either will be disappointed.

\subsection{A Second Modality: Keystroke Dynamics}
\label{sec:keystroke}
Nothing in the derivation is specific to eye movements: it needs a fixed-length vector, two sessions and an angular score. To test that directly we applied the law, unchanged, to the keystroke-dynamics benchmark of Killourhy and Maxion \cite{Killourhy2009}, in which 51 subjects typed the same password 50 times in each of eight sessions on separate days, described by 31 hand-crafted timing features (hold times and down--down and up--down latencies). This is the setting of Section~\ref{sec:plos}---hand-crafted features of a behavioural modality under cosine scoring---in a modality unrelated to ours. The template of a subject in a session is the mean of that session's 50 repetitions; features are $z$-scored over the pooled templates; session~1 is the enrollment template and session~2 or session~8 the probe.

Table~\ref{tab:keystroke} gives the result. With no fitted parameter, the two-normal form predicts $5.87\%$ against a measured $5.88\%$ one day later, and $18.33\%$ against $17.29\%$ at the longest interval; the equal-spread law with $\hat\rho$ and $\widehat N_{\mathrm{eff}}$ gives $5.17\%$ and $17.99\%$. The mixture form \eqref{eq:mixture}, evaluated at per-subject persistences estimated from other session pairs the same number of days apart---so that no information from the tested comparison enters---gives $5.76\%$ and $19.44\%$. The spread of persistence across subjects that it uses is itself informative: it grows from $0.08$ at a one-day interval to $0.22$ at the longest, so subjects diverge from one another as time passes, which is what Section~\ref{sec:subjects} says will make the law at the mean persistence progressively optimistic.

Three further readings are given in Section~S15 of the Supplementary Material. The 31 features behave like $6.9$ independent ones by the impostor spread and $3.5$ by the intercorrelation; rank-normalizing the features brings the mean excess kurtosis from $6.6$ to $-0.2$ and the law continues to hold; and averaging more repetitions raises the persistence less than Spearman--Brown predicts, because repetitions within a session share session-level variation.

\begin{table}[t]
\centering
\caption{The law on keystroke dynamics \cite{Killourhy2009}: 51 subjects, 31 timing features, enrollment in session~1 and probe in session~2 (one day later) or session~8. Predictions are unfitted: the equal-spread law with $\hat\rho$ and $\widehat N_{\mathrm{eff}}$ read off the scores, the two-normal form, and the mixture form \eqref{eq:mixture} at per-subject persistences estimated from other session pairs of the same interval.}
\label{tab:keystroke}
\footnotesize
\setlength{\tabcolsep}{4pt}
\begin{tabular}{@{}llrrrr@{}}
\toprule
Features & Probe & EER & law & two-normal & mixture \\
\midrule
raw & session 2 & 5.88 & 5.17 & 5.87 & 5.76 \\
raw & session 8 & 17.29 & 17.99 & 18.33 & 19.44 \\
rank-normalized & session 2 & 7.96 & 8.44 & 8.27 & 5.98 \\
rank-normalized & session 8 & 18.71 & 23.42 & 21.16 & 22.32 \\
\bottomrule
\end{tabular}
\end{table}

\subsection{A Third Modality, in Which Subjects Differ: the ECG}
\label{sec:ecg}
Keystroke dynamics is a modality in which the law's simple forms suffice. The ECG is not, and that makes it the sharper test of Section~\ref{sec:subjects}. We used the ECG-ID database \cite{Lugovaya2005,Goldberger2000}: 90 subjects, 20-second lead-I recordings at 500\,Hz, two to twenty records per subject taken from one day to six months apart. R peaks are located by the usual derivative--square--smooth rule, each beat is taken from 250\,ms before to 400\,ms after the peak and standardized, and the median beat of a record becomes its template, so a record is described by the 325 samples of one beat. Those samples are the features. They are strongly intercorrelated by construction, which also makes the database a test of \eqref{eq:neff}. Record 1 is the enrollment template and record 2 the probe, for the 41 subjects with a third record available.

The measured EER is $14.6\%$. The equal-spread law at $\hat\rho=0.89$ and $\widehat N_{\mathrm{eff}}=9.1$ predicts $2.0\%$: optimistic by a factor of seven, the largest failure of that form anywhere in this paper. The diagnostic of Section~\ref{sec:subjects} says why before any modelling: the genuine Fisher-$z$ spread is $2.1$ times the impostor spread, against $1.3$ for the embeddings of Section~\ref{sec:ekyt-scores}. The genuine scores are correspondingly spread out---their quartiles are $0.62$, $0.89$ and $0.95$---so a single $\rho$ cannot describe this population. The two-normal form, which absorbs the excess into one wider normal, recovers most of the gap at $13.1\%$. The mixture form \eqref{eq:mixture}, evaluated at per-subject persistences $\rho_i$ estimated from each subject's other record pairs so that the tested comparison contributes nothing, predicts $14.6\%$ (with 41 genuine comparisons, the EER resolves only to $2.4$ points). Those $\rho_i$ have mean $0.70$ and standard deviation $0.22$, and range down to $0.18$: the ECG population contains subjects whose beat is reproducible months later and subjects whose beat is barely reproducible at all, and it is the second group that sets the error rate.

The intercorrelation behaves as the law says as well. The 325 samples of a beat have a mean squared correlation of $0.14$, so \eqref{eq:neff} gives $7.2$ effective features and the impostor spread gives $9.1$. Three hundred and twenty-five features again deliver fewer than ten effective ones, as hand-crafted feature sets did in Section~\ref{sec:plos} and keystroke features in Section~\ref{sec:keystroke}; 128-dimensional embeddings reach only about $13$--$17$ (Section~\ref{sec:ekyt-saturation}). Adding samples to a beat template, like adding dimensions to an embedding, buys almost nothing once the samples are correlated.

\subsection{A Fourth Modality, in Which Recordings Differ: Faces}
\label{sec:face}
In the ECG, persistence varies between people. In face photographs it varies between recordings: two pictures of the same person differ in pose, illumination, expression and image quality, and the further apart in time they were taken, the more the face itself has changed. We used AgeDB \cite{Moschoglou2017AgeDB}, 16\,488 photographs of 567 labelled people (568 in the original description) collected from the web, each labelled with the person's age when it was taken. Faces were detected with SCRFD \cite{Guo2022SCRFD}, aligned to $112\times112$ pixels, and embedded by a 512-dimensional ArcFace network \cite{Deng2019ArcFace} (the InsightFace model \texttt{w600k\_r50}, trained on WebFace600K \cite{Zhu2021WebFace}). On LFW \cite{Huang2007LFW} the same pipeline reaches the published 10-fold verification accuracy, $99.85\pm0.20\%$, so the embeddings are sound; LFW has no repeated sessions and serves only as that check. In AgeDB the difference in age between two photographs plays the role of the interval between sessions. We formed five age-gap bins (3--5, 6--10, 11--20, 21--30 and more than 30 years); in each, every person with a qualifying pair contributes the pair whose gap is closest to the bin centre, with the younger photograph as enrollment and the older as probe. A 0--2-year bin was excluded because many of its pairs are near-duplicate photographs, which doubles its genuine Fisher-$z$ spread ($0.35$ against $0.16$--$0.21$ in the other bins). The ArcFace embeddings are already almost centred (impostor Fisher-$z$ mean $0.01$); centring on each bin's own mean, or on the mean of an independent set of LFW photographs, changes no EER by more than $0.4$ points, and the uncentred embeddings are used throughout.

\emph{The a-priori route fails.} Feature-level persistence falls with the age gap as expected, from a mean KCC of $0.76$ to $0.67$ (mean ICC $0.53$ to $0.36$; Table~\ref{tab:face}), and the 512 dimensions are only weakly intercorrelated, so \eqref{eq:neff} gives $\Neff=191$--$199$. The route of Section~\ref{sec:ekyt}---the law at the mean KCC and the $\Neff$ of \eqref{eq:neff}, with no score used---then predicts $0.002$--$0.40\%$ against a measured $3.2$--$6.2\%$, optimistic by one to three orders of magnitude. Neither the persistence nor the feature count is at fault: the genuine scores are centred where the features say they should be (median $0.56$ at 3--5 years, against a mean ICC of $0.53$), and the impostor spread confirms an effective feature count above $200$. What the route misses is the spread. The genuine Fisher-$z$ scores have a standard deviation of $0.16$--$0.21$, against $0.06$ for the impostor scores, a ratio of $2.6$--$3.2$ (embeddings of Section~\ref{sec:ekyt-scores}: $1.3$; ECG: $2.1$). Unlike in the ECG, the excess is not a property of people. The mixture form \eqref{eq:mixture} with per-subject persistences $\rho_i$, each estimated from that person's other pairs in the same bin, predicts only $0.01$--$1.7\%$: those $\rho_i$ have a standard deviation of $0.07$--$0.09$, too small to account for the genuine spread. Most of the spread lies between photographs of the same person.

\emph{The general form, on held-out subjects.} This is the case of Section~\ref{sec:subjects} in which persistence varies across comparisons rather than across people, and it allows a held-out test. In each of 20 random halvings of the subjects, the calibration half at the full 512 dimensions supplies the mean $m$ and spread $\tau$ of $\atanh\rho_c$, with $\tau^2$ the excess of the genuine over the impostor Fisher-$z$ variance, and the comparison-level persistences $\rho_c$, obtained by shrinking its genuine Fisher-$z$ scores toward $m$ as described in Section~\ref{sec:subjects}. The test half supplies the measured EER, on the full embedding and on random subsets of $K=4$--$256$ dimensions (10 subsets per size); for each subset $\Neff$ comes from the impostor spread \eqref{eq:neff-imp} of the same dimensions in the calibration half, the route used for every modality in Section~\ref{sec:modalities}. Persistence is assumed not to depend on which dimensions are used, so $m$ and $\tau$ are estimated once.

\begin{figure*}[t]
\centering
\includegraphics[width=0.9\textwidth]{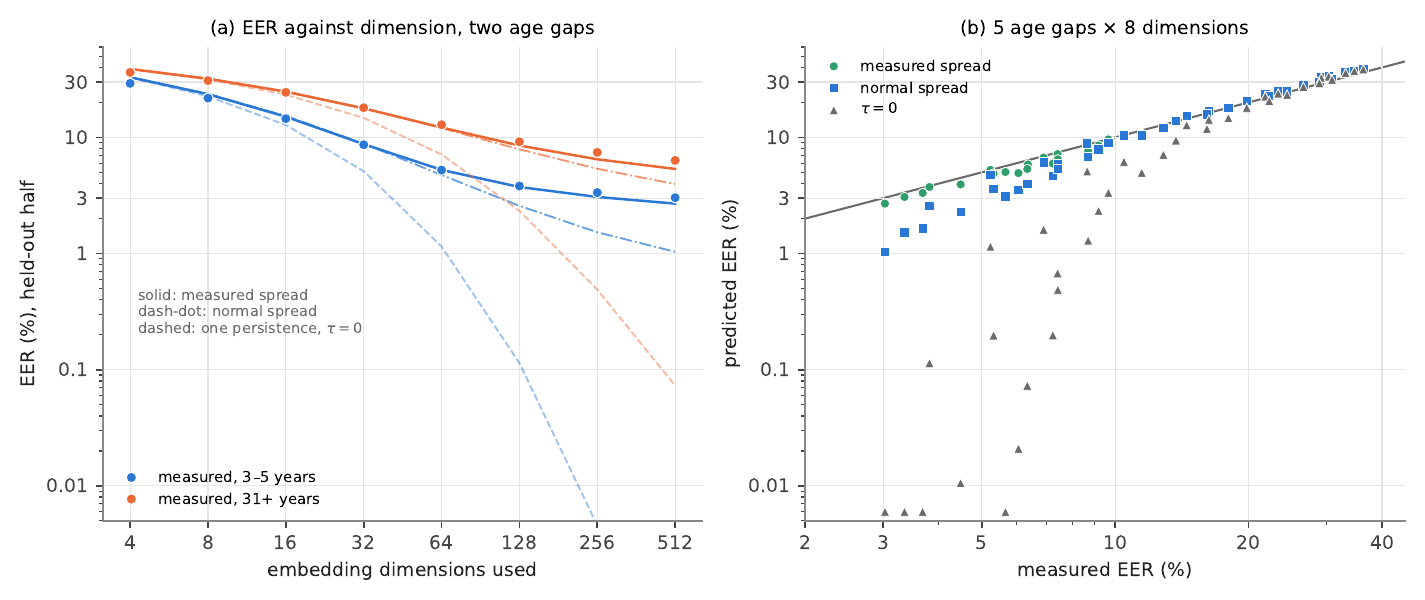}
\caption{Face embeddings (AgeDB), held-out halves, mean over 20 subject halvings and 10 random subsets of dimensions per size; $\tau$ and $\Neff$ from the impostor spread of the calibration half. (a) Measured EER (markers) against the number of embedding dimensions for the shortest and longest age gaps, with the mixture \eqref{eq:mixture} over the measured comparison-level persistences (solid), the general form \eqref{eq:unified} with a normal spread (dash-dot) and the law with one persistence, $\tau=0$ (dashed). (b) Predicted against measured EER for all five age gaps and eight dimension counts. Both axes logarithmic.}
\label{fig:face}
\end{figure*}

The calibration halves give $\tau=0.15$--$0.20$ and $\tau\sqrt{\Neff-2}=2.4$--$3.1$ in every bin: faces sit close to the spread-limited regime of \eqref{eq:unified}. Fig.~\ref{fig:face}a shows what that means. The law with one persistence ($\tau=0$) follows the measured EER up to about 16 dimensions and then falls away, predicting less than $0.1\%$ at 512 dimensions against a measured $3.0$--$6.3\%$; the measured EER instead levels off, well above the normal-spread floor \eqref{eq:floor} ($0.07$--$0.5\%$). The general form with a normal spread, \eqref{eq:unified}, reproduces the levelling but stays far too low at full dimension ($1.0$--$4.0\%$), because the lower tail of $\rho_c$ is heavier than normal---the situation in which Section~\ref{sec:subjects} says two moments are not enough. The mixture \eqref{eq:mixture} over the measured $\rho_c$ tracks the data at every gap and dimension (Table~\ref{tab:face}). Over all 40 combinations of gap and dimension it explains $R^2=0.982$ of the measured EER with a median absolute error of $0.59$ points, against $0.967$ and $1.62$ for the normal spread and $0.806$ and $3.69$ for $\tau=0$ (Fig.~\ref{fig:face}b). Two departures remain. At 4 dimensions all three forms over-predict by about 3 points; there $\Neff\approx4$ and Fisher's normal approximation is coarse (Section~\ref{sec:exact}). At 512 dimensions the mixture is slightly optimistic, $2.7$--$5.4\%$ against $3.0$--$6.3\%$: the impostor spread of the calibration half implies about 250 effective features, while the intercorrelation \eqref{eq:neff}, estimated on these halves of about 280 subjects, implies about 160 (sampling noise inflates $\overline{r^2}$; all pairs give $191$--$199$). With $\Neff$ from \eqref{eq:neff} the mixture predicts $2.9$--$6.2\%$ at 512 dimensions and has a median error of $0.39$ points over the 40 combinations ($R^2=0.977$), so the choice of estimator matters by about one point at most.

Faces and eye movements are therefore limited by different terms of the same law, although their persistence spreads are similar: measured the same way, the eye-movement embeddings have $\tau\approx0.21$ (Section~\ref{sec:modalities}). The difference is the sampling spread. With about 16 effective features it is $0.26$ for the eye-movement embeddings, comparable to $\tau$; with about 250 it is $0.06$ for ArcFace, and the error is set by how much persistence varies from one pair of photographs to the next. Accordingly, going from 128 to 512 face dimensions lowers the EER at the shortest gap only from $3.8$ to $3.0\%$, where the law with $\tau=0$ would have it fall to almost nothing.

\begin{table*}[t]
\centering
\caption{The law on face embeddings (AgeDB \cite{Moschoglou2017AgeDB}, ArcFace, 512 dimensions), one pair per person in each age-gap bin. Left: all pairs; feature-level KCC and ICC between the younger and older embeddings, and the a-priori law (KCC route of Section~\ref{sec:ekyt}, $\Neff$ from \eqref{eq:neff}, no scores used). Right: held-out test, mean over 20 subject halvings; $\tau$ and $\Neff$ from the impostor spread of the calibration half, EER measured on the test half, predictions with one persistence ($\tau=0$), a normal spread \eqref{eq:unified} and the measured spread \eqref{eq:mixture}; last column: measured spread with $\Neff$ from \eqref{eq:neff} instead. EERs in percent.}
\label{tab:face}
\footnotesize
\setlength{\tabcolsep}{4.5pt}
\begin{tabular}{@{}lrrrrrrrrrrrr@{}}
\toprule
& \multicolumn{6}{c}{all pairs} & \multicolumn{6}{c}{held-out halves, 512 dimensions} \\
\cmidrule(lr){2-7}\cmidrule(l){8-13}
Gap (years) & mean gap & pairs & KCC & ICC & a-priori law & EER & $\tau$ & EER & $\tau=0$ & normal spread & measured spread & \eqref{eq:neff} \\
\midrule
3--5    & 4.0  & 566 & 0.756 & 0.528 & 0.002 & 3.18 & 0.19 & 3.03 & 0.00 & 1.04 & 2.69 & 2.90 \\
6--10   & 8.0  & 565 & 0.743 & 0.501 & 0.006 & 3.89 & 0.20 & 3.68 & 0.00 & 1.65 & 3.32 & 3.73 \\
11--20  & 15.3 & 563 & 0.713 & 0.438 & 0.05  & 6.04 & 0.19 & 5.66 & 0.01 & 3.10 & 5.04 & 5.57 \\
21--30  & 25.4 & 553 & 0.692 & 0.395 & 0.15  & 5.97 & 0.17 & 6.05 & 0.02 & 3.51 & 4.93 & 5.64 \\
31+     & 37.6 & 517 & 0.673 & 0.356 & 0.40  & 6.19 & 0.15 & 6.34 & 0.07 & 3.96 & 5.35 & 6.15 \\
\bottomrule
\end{tabular}
\end{table*}

\subsection{One Law, Four Modalities}
\label{sec:modalities}
\begin{figure*}[t]
\centering
\includegraphics[width=\textwidth]{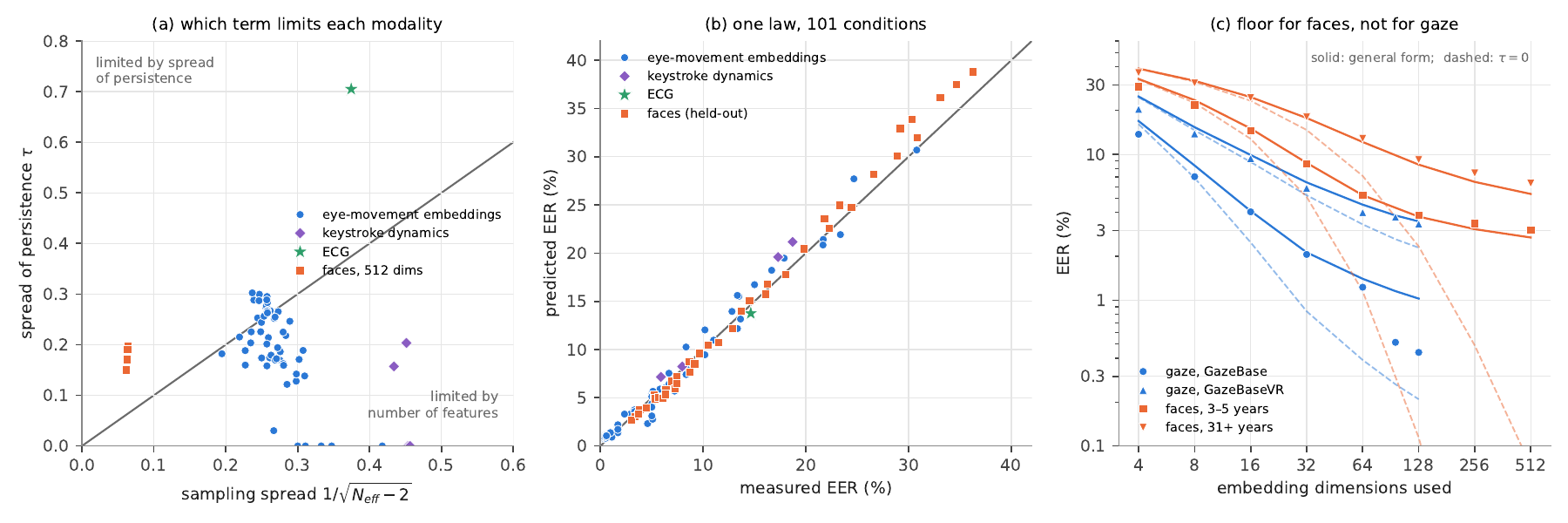}
\caption{One law, four modalities; $\Neff$ and $\tau$ from the impostor spread throughout. (a) Spread of persistence $\tau$ against the sampling spread $1/\sqrt{\Neff-2}$ for every condition; above the diagonal the spread of persistence is the larger term, below it the sampling spread. (b) EER predicted by the general form with the measured spread \eqref{eq:mixture} against measured EER, 101 points; gaze, keystroke and ECG in-sample, faces held-out. (c) EER against the number of embedding dimensions for gaze (unmanipulated GazeBase and GazeBaseVR, random subsets of Section~\ref{sec:ekyt-saturation}) and faces (shortest and longest age gap): markers measured, solid the general form, dashed $\tau=0$.}
\label{fig:modalities}
\end{figure*}

The general form \eqref{eq:unified}, with \eqref{eq:mixture} over the measured spread, can now be applied to every real dataset in this paper in the same way. Throughout, $\Neff$ is read from the impostor spread \eqref{eq:neff-imp}, $m$ from the mean genuine Fisher-$z$ score less the mean impostor score, and $\tau^2$ from the excess of the genuine over the impostor variance; no parameter is fitted. For the eye-movement embeddings (the 56 conditions of Section~\ref{sec:ekyt-scores}), keystroke dynamics (four conditions) and the ECG, $m$ and $\tau$ come from the same scores whose EER is predicted, as for the two-normal form; for faces they come from held-out subjects (Section~\ref{sec:face}). Table~\ref{tab:modalities} and Fig.~\ref{fig:modalities} give the result.

\emph{Which term limits each modality.} The ratio of the two spreads in \eqref{eq:unified}, $\tau\sqrt{\Neff-2}=\tau/\sqrt v$, places each modality between the two regimes of Section~\ref{sec:subjects} (Fig.~\ref{fig:modalities}a). Keystroke dynamics lies well inside the feature-limited regime (ratio $0.18$, the median of four conditions, two with $\tau$ clipped at zero): the spread of persistence is small beside the sampling spread, and the error is set by the handful of effective features. The eye-movement embeddings sit where the two spreads are about equal (median ratio $0.83$; the 56 conditions lie on either side of one): both terms matter, which is why the two-normal form, with both spreads measured, was needed in Section~\ref{sec:ekyt-scores}. The ECG (ratio $1.9$) is mostly spread-limited and faces ($2.9$) are close to that limit, for different reasons: in the ECG persistence varies between people (Section~\ref{sec:ecg}), in faces between photographs of the same person (Section~\ref{sec:face}).

\emph{How well one law does.} Over all 101 points---56 gaze conditions, four keystroke conditions, the ECG and 40 face combinations of age gap and dimension---the general form with the measured spread explains $R^2=0.982$ of the measured EER, with a median absolute error of $0.58$ points; the law with one persistence ($\tau=0$) gives $0.870$ and $1.87$ (Fig.~\ref{fig:modalities}b). Within each modality the measured-spread form is the most accurate or level with the best (Table~\ref{tab:modalities}), with one exception. The raw keystroke latencies have a genuine Fisher-$z$ spread smaller than the impostor spread ($0.87$--$0.90$ of it), because their heavy tails widen the impostor distribution (Section~\ref{sec:sim-kcc}); $\tau^2$ is then clipped at zero, and the general form reduces to the law with one persistence at a sampling spread inflated by the tails. It predicts $7.2$ and $19.6\%$ against $5.9$ and $17.3\%$, where the two-normal form, which takes both spreads as measured, gives $5.9$ and $18.3\%$. Rank-normalizing the features, as Section~\ref{sec:sim-kcc} recommends, restores a genuine spread above the impostor spread, and the general form and the two-normal form then agree ($8.2$ and $21.2\%$ against $8.0$ and $18.7\%$ measured).

\emph{Baseline conditions.} Table~SVII of the Supplementary Material (Section~S16) sets the measured EER of the baseline condition of each modality beside the predictions: the unmanipulated recordings for gaze, the four keystroke conditions, the ECG, and faces at full dimension for every age gap. The general form is within $0.2$ points on unmanipulated GazeBaseVR ($3.47\%$ against $3.33\%$) and within one point on the ECG ($13.75\%$ against $14.63\%$; the subject mixture of Section~\ref{sec:ecg} gives $14.58\%$). On faces it is $0.3$--$1.1$ points low ($2.69$--$5.35\%$ against $3.03$--$6.34\%$), where the law with $\tau=0$ predicts less than $0.1\%$. On unmanipulated GazeBase it predicts $1.03\%$ against a measured $0.44\%$, below the resolution of one genuine comparison in 59.

\emph{Two modalities, one panel.} Fig.~\ref{fig:modalities}c shows the two regimes against the number of dimensions, using the random subsets of Section~\ref{sec:ekyt-saturation} for gaze (with $m$ and $\tau$ from all 128 dimensions) and those of Section~\ref{sec:face} for faces. The gaze EER keeps falling until the effective number of features saturates; the face EER levels off while $\Neff$ is still growing. The general form follows both. The law with $\tau=0$ is roughly right for gaze but misses faces by orders of magnitude. Over the 28 gaze combinations of condition and dimension the general form has a median error of $0.53$ points ($R^2=0.950$), against $1.16$ for $\tau=0$. The largest gaze departure is for unmanipulated GazeBase at 128 dimensions, $1.0\%$ predicted against $0.44\%$ measured, a measured EER below the resolution of one genuine comparison in 59 ($1.7\%$).

The answer to the question posed in the Introduction can therefore be read off a pair of score distributions. If $\tau\sqrt{\Neff-2}$ is well below one, the number of effective features limits the error; if it is about three or more, the spread of persistence does; between the two, both matter.

\begin{table*}[t]
\centering
\caption{One law, four modalities. $\tau$: spread of persistence on Fisher's scale; $\sqrt v=1/\sqrt{\Neff-2}$: sampling spread, both from the impostor route (medians over conditions; faces at 512 dimensions). Median absolute error (EER points) of four forms: one persistence ($\tau=0$), the two-normal form (both spreads measured), the general form with a normal spread \eqref{eq:unified}, and with the measured spread \eqref{eq:mixture}. Gaze, keystroke and ECG in-sample; faces held-out (two-normal form not evaluated).}
\label{tab:modalities}
\footnotesize
\setlength{\tabcolsep}{0pt}
\begin{tabular*}{\textwidth}{@{\extracolsep{\fill}}lrlrrrlrrrr@{}}
\toprule
& & & & & & & \multicolumn{4}{c}{median $|$error$|$ (points)} \\
\cmidrule{8-11}
& & variation & & & & & & two- & normal & measured \\
Modality & points & of $\rho_c$ & $\tau$ & $\sqrt v$ & $\tau/\sqrt v$ & limited by & $\tau=0$ & normal & spread & spread \\
\midrule
Keystroke & 4 & little & 0.08 & 0.45 & 0.18 & features & 1.31 & 0.67 & 1.79 & 1.79 \\
Eye movements & 56 & modest & 0.21 & 0.26 & 0.83 & both & 1.12 & 0.63 & 0.63 & 0.55 \\
ECG & 1 & between people & 0.71 & 0.38 & 1.88 & mostly spread & 10.71 & 1.58 & 1.58 & 0.88 \\
Faces (held-out) & 40 & between photos & 0.19 & 0.06 & 2.93 & near spread limit & 3.69 & --- & 1.62 & 0.59 \\
\midrule
All & 101 & & & & & & 1.87 & & & 0.58 \\
\bottomrule
\end{tabular*}
\end{table*}

\section{Design Implications}
\label{sec:design}
The law turns the planning questions of the Introduction into arithmetic. Table~\ref{tab:design} evaluates the finite-$N$ form of the law, with $\rho$ obtained from the KCC through \eqref{eq:kcc2rho}, for $N=50$, $100$ and $200$ features at three levels of KCC, first for truly independent features and then at two levels of intercorrelation; Fig.~S4 of the Supplementary Material (Section~S17) shows the same relations graphically.

\begin{table}[t]
\centering
\caption{Predicted EER (\%) for $N$ features of a given KCC, independent or with the intercorrelation stated ($\overline{r^2}=0.01$, i.e.\ all pairs at $0.10$, and $\overline{r^2}=0.056$, the value for the embeddings of \cite{Raju2026TBIOM}). Finite-$N$ form of \eqref{eq:law} with $\rho$ from \eqref{eq:kcc2rho}.}
\label{tab:design}
\small
\begin{tabular}{@{}llrrr@{}}
\toprule
KCC ($\rho$) & $N$ & independent & $\overline{|r|}\approx0.10$ & $\overline{|r|}\approx0.19$ \\
\midrule
0.7 (0.42) & 50 & 6.1 & 10.3 ($\Neff\,34$) & 21.8 ($\Neff\,13$) \\
 & 100 & 1.4 & 6.0 ($\Neff\,50$) & 20.1 ($\Neff\,15$) \\
 & 200 & 0.09 & 3.6 ($\Neff\,67$) & 19.2 ($\Neff\,17$) \\
0.8 (0.62) & 50 & 0.58 & 2.0 & 10.2 \\
 & 100 & 0.016 & 0.57 & 8.6 \\
 & 200 & $2\times10^{-5}$ & 0.17 & 7.8 \\
0.9 (0.81) & 50 & 0.003 & 0.06 & 2.3 \\
 & 100 & $\approx10^{-6}$ & 0.003 & 1.6 \\
 & 200 & $\approx0$ & $2\times10^{-4}$ & 1.3 \\
\bottomrule
\end{tabular}
\end{table}

Four conclusions follow.

\emph{Independent features are extraordinarily valuable.} With $\KCC=0.8$ ($\rho\approx0.62$), 43 independent features give $\EER=1\%$ and 75 give $0.1\%$; with $\KCC=0.9$ the numbers are 18 and 31. A hundred independent features at $\KCC=0.8$ would give $\EER\approx0.02\%$.

\emph{Intercorrelation caps what dimensions can buy.} At the intercorrelation of the embeddings in \cite{Raju2026TBIOM}, 50, 100 and 200 dimensions are worth $\Neff=13$, $15$ and $17$; the EER at $\KCC=0.8$ moves only from $10\%$ to $8\%$. Doubling the width of an embedding layer is nearly worthless unless the added dimensions are decorrelated from the existing ones. Section~\ref{sec:ekyt-saturation} measures this directly on the embeddings of that study: going from 64 to 128 dimensions adds $7$--$14\%$ to $\Neff$, and the EER flattens. A second limit appears there as well---the spread of persistence across comparisons, which sets a floor that added dimensions cannot lower (the fourth conclusion below).

\emph{The three levers trade off through one number.} Everything enters through $\dprime=\sqrt{\Neff}\,\atanh\rho$. Doubling $\Neff$ gains a factor $1.41$ in $\dprime$. Raising KCC from $0.8$ to $0.9$ gains a factor $1.6$ ($\atanh0.81/\atanh0.62$). Reducing the mean absolute intercorrelation of a 128-dimensional embedding from $0.19$ to $0.10$ raises $\Neff$ from $16$ to $27$ if the spread of the correlations stays at $0.14$, to $43$ if it shrinks in proportion, and to $56$ if it vanishes---a factor $1.3$--$1.9$ in $\dprime$. For learned embeddings that already sit at $\Neff\approx16$, decorrelation---a whitening layer, or a decorrelation penalty in the training loss---is the cheapest lever, and Section~\ref{sec:whitening} says how to assess it: through the mean canonical persistence rather than the raw per-dimension KCC.

\emph{The spread of persistence sets a floor, and decides which lever works.} The three levers above assume one persistence for every comparison. The general form \eqref{eq:unified} adds a fourth, the uniformity of persistence across comparisons, and makes the value of the other three depend on it. The argument of $\Phi$ in \eqref{eq:unified}, $m/(\sqrt{\tau^2+v}+\sqrt v)$, rises toward $m/\tau$, the floor \eqref{eq:floor}, as $\Neff$ grows. With $x=\tau\sqrt{\Neff-2}$, the fraction of that limit already reached is $1/(\sqrt{1+1/x^2}+1/x)$: $0.41$ at $x=1$, $0.72$ at $x=3$ and $0.90$ at $x=10$. The rule for a designer follows. When $x\ll1$, as for keystroke dynamics ($x\approx0.2$; Table~\ref{tab:modalities}), the error is limited by the number of effective features, and adding or decorrelating features pays as described above. When $x\gtrsim3$, most of the attainable separation has already been reached; more or less correlated features buy little, and only raising persistence or making it more uniform helps. Between these, both levers act: at $x=1$, doubling $\Neff-2$ still raises the argument of $\Phi$ by a quarter. The ECG ($x=1.9$, $60\%$ of the limit) and faces ($2.9$, $71\%$) lie toward the spread-limited end, where persistence is the better lever---for faces, anything that reduces the variation between recordings; for the ECG, identifying and treating the subjects of low persistence. The eye-movement embeddings, at $x\approx0.8$, sit where both levers act. With $\tau\approx0.2$, the value found for both gaze and faces, $x$ reaches $1$ at $\Neff\approx27$ and $3$ at $\Neff\approx227$, so a 512-dimensional face embedding with about 250 effective features is at the point beyond which more dimensions pay little. Because $\tau$ and $\Neff$ are read from the genuine and impostor scores of a pilot set (Section~\ref{sec:subjects}), the rule can be applied before a system is built out.

Which form of the law to use follows from the same scores, and Table~SVIII of the Supplementary Material collects the cases met in this paper. The diagnostic is the ratio of the genuine to the impostor Fisher-$z$ spread, $\sigma_g/\sigma_i$, and, when it exceeds one, whether the normal-spread and measured-spread forms agree.

\section{Assumptions, Limitations, and a Validation Protocol}
\label{sec:limitations}

\subsection{What the Derivation Assumes}
The derivation is exact for the model of Section~\ref{sec:model}---standardized, independent, normally distributed features scored by cosine similarity, with session~2 differing from session~1 only by fresh noise---and the closed form is its large-$N$ limit. Each departure has been examined above and enters in a known way. Unequal persistence: use the mean (Section~\ref{sec:unequal}); the residual is a slight optimism. Persistence that varies across comparisons---between subjects or between recordings: use \eqref{eq:unified}, or the mixture form \eqref{eq:mixture} when the spread has a heavy lower tail (Section~\ref{sec:subjects}); the law at the mean persistence is optimistic, by a factor of two or more at realistic spreads and by a factor of seven on the ECG data of Section~\ref{sec:ecg}. The ratio of the genuine to the impostor spread in Fisher-$z$ units is the diagnostic. Intercorrelation: use $\Neff$, a first-order rule that is pessimistic when one common factor dominates (Section~\ref{sec:neff}); or whiten and use the canonical spectrum, which is accurate (Section~\ref{sec:whitening}). Session-level changes: use the score-distribution estimates $\hat\rho$ and $\widehat N_{\mathrm{eff}}$ (Section~\ref{sec:session}). Non-normal features: the KCC mapping \eqref{eq:kcc2rho} is blind to monotone distortion, to which the cosine is not, so it may be used while the typical dimension has excess kurtosis of about $1$ or less and otherwise after rank-normalizing the features (Section~\ref{sec:sim-kcc}). Uncentred embeddings: if the embedding vectors share a large common mean, impostor cosine scores are not centred at zero and the law does not apply as written; centring on the training-set mean restores the setting of the derivation, and the impostor median is the diagnostic. Averaged enrollment: feed the law the persistence of the centroids (Section~\ref{sec:averaging}).

The general form has limitations of its own. First, the spread $\tau$ cannot be computed from feature-level persistence: it needs genuine scores from pilot subjects, and when it is ignored the a-priori route can fail by orders of magnitude, as it does on faces (Section~\ref{sec:face}). Second, only the face results and the ECG mixture are held-out tests; for the eye-movement embeddings and keystroke dynamics $m$ and $\tau$ come from the same scores whose EER is predicted, as for the two-normal form, so those comparisons check consistency rather than prediction, and the 59--60 test subjects of the gaze data are too few to split. Third, $\tau^2$ is a difference of two variances. Heavy-tailed features can make the genuine spread smaller than the impostor spread, $\tau^2$ is then clipped at zero, and the two-normal form does better (raw keystroke latencies, Section~\ref{sec:modalities}); rank-normalizing removes the problem (Section~\ref{sec:sim-kcc}). Session noise shared across features also inflates the genuine variance, so $\tau$ is an upper bound unless that cause can be excluded (Section~\ref{sec:subjects}). Fourth, the general form reads $\Neff$ from the impostor spread \eqref{eq:neff-imp}, while \eqref{eq:neff} remains the a-priori approximation for design. The two can differ: for keystroke dynamics only the impostor route works ($6.9$ against $3.5$ effective features), and for faces the impostor route gives about 250 effective features against 160, which changes the full-dimension prediction by up to $0.8$ points. Fifth, $m$ and $\tau$ are assumed not to depend on which dimensions are used. That held for gaze and faces from 8 dimensions up, but at 4 dimensions every form over-predicts the EER by 3--5 points; there $\Neff\approx4$, Fisher's normal approximation is coarse, and the exact finite-$N$ density of Section~\ref{sec:exact} has not yet been carried over to the general form.

The condition-level comparisons of Section~\ref{sec:realdata} are consistent with the law but do not test it sharply, because they rest on median ICCs, on condition-level summaries, and on 59--60 test subjects. The staged protocol below tests it directly.

\subsection{A Staged Validation Protocol}
The protocol has five stages, set out in full in Section~S19 of the Supplementary Material. Stage~1 asks whether the two-normal Fisher-$z$ picture holds for embeddings (Section~\ref{sec:ekyt-scores}: after centring, yes, in the two-normal form). Stage~2 asks which persistence index predicts $\hat\rho$ (Section~\ref{sec:ekyt-index}: all three do, equally well). Stage~3 asks whether intercorrelation sets $\Neff$ (Section~\ref{sec:ekyt-saturation}: yes, and the EER floor that remains is set by persistence varying across comparisons). Stage~4, not yet carried out, tests whether whitening estimated on training-round embeddings delivers the canonical prediction of Section~\ref{sec:whitening}. Stage~5, also not yet carried out, recomputes the 29 subsets of \cite{Friedman2017PLoS} with the mean ICC and the participation ratio of the retained components, to separate effective dimension from persistence that varies across subjects (Section~\ref{sec:plos}).

\section{Conclusion}
The relation between temporal persistence and biometric performance that our group has documented empirically for a decade has a simple theory. Cosine similarity between two feature vectors is a correlation coefficient; genuine comparisons have population correlation equal to the persistence of the features, impostor comparisons have zero; and Fisher's transform reduces the verification problem to two normals. When every comparison shares one persistence, the result is $\EER=\Phi(-\tfrac12\sqrt N\,\atanh\rho)$, with inverse $N=(2q/\atanh\rho)^2$. It reproduces with no fitted parameter the feature-count tables and score-distribution findings of our earlier simulation studies, explains the published straight-line relations as local views of a curve whose shape is set by the effective number of independent features, and predicts the performance of deep eye-movement embeddings on GazeBase from their intercorrelation alone.

Real comparisons do not share one persistence. Allowing for its spread $\tau$ gives the general form of the law, with a sampling term that shrinks as the effective number of features grows and a spread term that does not, and, for a normal spread, with a floor $\Phi(-m/\tau)$ that no number of features can lower. Nothing in it is specific to eye movements. Read from the genuine and impostor scores, it matches the EER of eye-movement embeddings, keystroke dynamics, the ECG and face embeddings over 101 conditions with $R^2=0.98$ and a median error of $0.6$ points, as a held-out test for faces and as a consistency check on the same scores for the other three. The four modalities span its range: keystroke dynamics is limited by its handful of effective features; the ECG and faces lie toward the limit set by the spread of persistence---between people in the ECG, between photographs in faces, where the simple law is optimistic by one to three orders of magnitude; and eye movements sit between. Its message for design therefore has two parts. While $\tau\sqrt{\Neff-2}$ is well below one, persistence and the number of effective features trade off as the simple law says: below a test--retest correlation of about $0.6$ (KCC about $0.8$) no realistic number of correlated features will reach low error rates, above $0.9$ (KCC about $0.95$) a dozen or two suffice, and intercorrelation places a ceiling on the value of added dimensions that only decorrelation can lift. Once $\tau\sqrt{\Neff-2}$ exceeds about three, added or decorrelated features buy little, and only higher or more uniform persistence lowers the error.

\bibliographystyle{unsrt}
\bibliography{ms.bib}

\end{document}


\title{Supplementary Material for \textit{How Many Persistent Features Does a Biometric Need? A Law Linking Equal Error Rate to Feature Count and Persistence}}
\renewcommand{\shorttitle}{A Law Linking Equal Error Rate to Feature Count and Persistence}

\author{
    Oleg Komogortsev, Mehedi Hasan Raju\\
    Texas State University, San Marcos, Texas, USA\\
    {\small\texttt{ok@txstate.edu, m.raju@txstate.edu}}
}
\date{}

\maketitle

\noindent Sections, equations and tables numbered with the prefix S belong to this document; all other numbers refer to the main paper.

\section{Kendall's $W$ for Two Raters}
\label{sup:kcc}
For $m$ rankings of $n$ objects, Kendall and Babington Smith \cite{Kendall1939} define $W=12S/[m^2(n^3-n)]$, where $S$ is the sum of squared deviations of the rank sums from their mean, and show that the average Spearman correlation among the $\binom m2$ pairs of rankings is $\bar r_s=(mW-1)/(m-1)$. With $m=2$ this is $r_s=2W-1$, i.e.\ (4). With ties, the tie-corrected $W$ and $r_s$ differ slightly; function \texttt{kcc2.m} computes both, and on simulated normal data the identity holds to four decimal places. Pearson's relation $r_s=(6/\pi)\arcsin(\rho/2)$ for a bivariate normal population \cite{Pearson1907} is inverted to give (12); the maximum difference between $r_s$ and $\rho$ over $[0,1]$ is $0.018$ at $\rho\approx0.59$.

\section{Variance of an Impostor Inner Product}
\label{sup:neff}
Let $\mathbf x$ and $\mathbf y$ be independent, zero-mean, with covariance $\mathbf R$. Then $\E[(\mathbf x^{\!\top}\mathbf y)^2]=\E[\mathbf x^{\!\top}\mathbf y\mathbf y^{\!\top}\mathbf x]=\E[\mathbf x^{\!\top}\mathbf R\mathbf x]=\tr(\mathbf R\,\E[\mathbf x\mathbf x^{\!\top}])=\tr(\mathbf R^2)$. For a correlation matrix $\tr(\mathbf R^2)=N+\sum_{j\ne k}r_{jk}^2=N[1+(N-1)\overline{r^2}]$, which gives (18). For a genuine pair with cross-covariance $\mathbf C$ the same calculation gives $\Var(\mathbf x_1^{\!\top}\mathbf x_2)=\tr(\mathbf R^2)+\tr(\mathbf C^2)$; the normalization in (6) removes part of this variance (for independent features, from $(1+\rho^2)N$ to $(1-\rho^2)^2N$), which is why the Fisher-$z$ route, rather than a direct moment calculation, gives the clean result.

\section{Exact Finite-$N$ Computation}
\label{sup:exact}
The density of the sample correlation coefficient with $\nu$ degrees of freedom and population correlation $\rho$ is \cite{Fisher1915,Hotelling1953}
\begin{multline}
  f_\nu(r;\rho)=\frac{(\nu-1)\,\Gamma(\nu)}{\sqrt{2\pi}\,\Gamma(\nu+\tfrac12)}\,
  (1-\rho^2)^{\nu/2}(1-r^2)^{(\nu-3)/2}\\
  \times(1-\rho r)^{-\nu+1/2}\;
  {}_2F_1\!\left(\tfrac12,\tfrac12;\nu+\tfrac12;\tfrac{1+\rho r}{2}\right),
  \label{eq:fisher-density}
\end{multline}
with ${}_2F_1$ the Gauss hypergeometric function. Genuine scores follow $f_N(\cdot;\rho)$ and impostor scores $f_N(\cdot;0)$, for which $r^2\sim\mathrm{Beta}(\tfrac12,\tfrac{N-1}{2})$.

Function \texttt{eerExact.m} evaluates \eqref{eq:fisher-density} after the substitution $r=\sin\theta$, which turns the factor $(1-r^2)^{(\nu-3)/2}\,dr$ into the bounded $\cos^{\nu-2}\theta\,d\theta$ and so removes the endpoint singularity of the density at $\nu=2$. It uses a grid of $40{,}001$ points uniform in $\theta\in[-\pi/2,\pi/2]$, computing ${}_2F_1(\tfrac12,\tfrac12;\nu+\tfrac12;z)$ by its power series (which converges for the arguments $z=(1+\rho r)/2<1$ that arise), integrates the genuine and impostor densities by the trapezoidal rule, and locates the FRR--FAR crossing by linear interpolation. Table~\ref{tab:exact} shows agreement with simulation to within simulation noise (about $0.1$ EER point) at every $N$ tested.

\section{Decorrelation and the Canonical Persistence Spectrum}
\label{sup:whitening}
The first-order rule describes cosine scoring of the correlated features as they are. What if the features are first decorrelated? Let $\boldsymbol\Sigma$ be the within-session covariance matrix of the features and $\mathbf C$ the cross-session covariance, $C_{jk}=\mathrm{cov}(x_{j1},x_{k2})$; the diagonal of $\mathbf C$ holds the feature persistences. Zero-phase (ZCA) whitening \cite{Kessy2018}, $\mathbf y=\boldsymbol\Sigma^{-1/2}\mathbf x$, produces $N$ uncorrelated unit-variance components whose cross-session covariance is
\begin{equation}
  \mathbf M=\boldsymbol\Sigma^{-1/2}\,\mathbf C\,\boldsymbol\Sigma^{-1/2}.
\end{equation}
Rotating to the eigenvectors of $\mathbf M$ (symmetrized, since a sample $\mathbf C$ is not exactly symmetric) gives $N$ components that are uncorrelated within a session and have cross-session correlations $\lambda_1,\dots,\lambda_N$, the eigenvalues of $\mathbf M$, which lie in $[-1,1]$ and are non-negative under the model of Section~III-A. We call these the \emph{canonical persistences} (canonical ICCs) of the feature set: they are the persistences of the decorrelated components and are invariant to any invertible linear transformation of the features. Because cosine similarity is invariant to rotations, cosine scoring of the whitened features is cosine scoring of $N$ independent features with persistences $\lambda_k$, so by Section~IV-A
\begin{equation}
  \EER_{\mathrm{whitened}}\approx\Phi\!\left(-\tfrac12\sqrt N\,\atanh\bar\lambda\right),\qquad
  \bar\lambda=\frac{\tr(\boldsymbol\Sigma^{-1}\mathbf C)}{N}.
  \label{eq:canonical}
\end{equation}
Section~V-H shows that this is accurate to well under one EER point across all correlation structures tested.

Equation~\eqref{eq:canonical} makes precise an observation from our earlier work: decorrelating a feature set changes its ICCs. The mean canonical persistence equals the mean raw persistence when the trait and the noise share the same correlation structure, is lower when the correlation is carried by the traits (decorrelation removes shared, persistent information), and is higher when the correlation is carried by the session noise (decorrelation removes shared noise). Two further remarks follow. PCA as used in \cite{Friedman2017PLoS} is a rotation, so it does not change cosine similarity at all except through the components it discards; only a rescaling step (whitening) can change $\Neff$. And a system that whitens its features is described by \eqref{eq:canonical} with the full $N$, so the canonical spectrum is the right object for assessing what a feature set \emph{could} deliver, while (18) describes what raw cosine scoring \emph{does} deliver.

\section{Averaged Enrollment}
\label{sup:averaging}
For averaging on both sides, the Spearman--Brown value of Section~IV-F applies; for one-sided averaging the persistence of the centroid is $\rho/\sqrt{\rho+(1-\rho)/k}$ instead. The script for the session changes of Section~\ref{sup:session} (\texttt{exp7\_session\_effects.m}) also tests the two-sided value: averaging $k=1,3,12$ sequences on both sides of an $N=16$, $\rho=0.5$ set gave simulated EERs of $14.2$, $3.1$ and $0.10\%$ against $14.4$, $3.0$ and $0.09\%$ predicted.

\section{Exact Versus Approximate EER}
\label{sup:exact-table}
Table~\ref{tab:exact} gives the comparison discussed in Section~V of the main paper.

\begin{table}[t]
\centering
\caption{EER (\%) from simulation (5,000 subjects), from the exact expression \eqref{eq:fisher-density}, from the closed form (8), and from its finite-$N$ form.}
\label{tab:exact}
\begin{tabular}{@{}rrrrrr@{}}
\toprule
$N$ & ICC & simulated & exact & law & law, $N\!-\!1$ \\
\midrule
2 & 0.75 & 28.5 & 28.3 & 24.6 & 31.3 \\
2 & 0.95 & 16.5 & 16.3 & 9.8 & 18.0 \\
4 & 0.75 & 18.9 & 18.8 & 16.5 & 20.0 \\
4 & 0.95 & 5.8 & 5.8 & 3.3 & 5.6 \\
8 & 0.55 & 20.0 & 20.1 & 19.1 & 20.7 \\
8 & 0.85 & 5.0 & 4.9 & 3.8 & 4.8 \\
16 & 0.75 & 2.96 & 3.04 & 2.58 & 2.98 \\
32 & 0.55 & 4.27 & 4.26 & 4.01 & 4.26 \\
32 & 0.85 & 0.020 & 0.032 & 0.019 & 0.024 \\
\bottomrule
\end{tabular}
\end{table}

\section{Features Required for a Target EER}
\label{sup:target}
\begin{figure}[t]
\centering
\includegraphics[width=0.8\columnwidth]{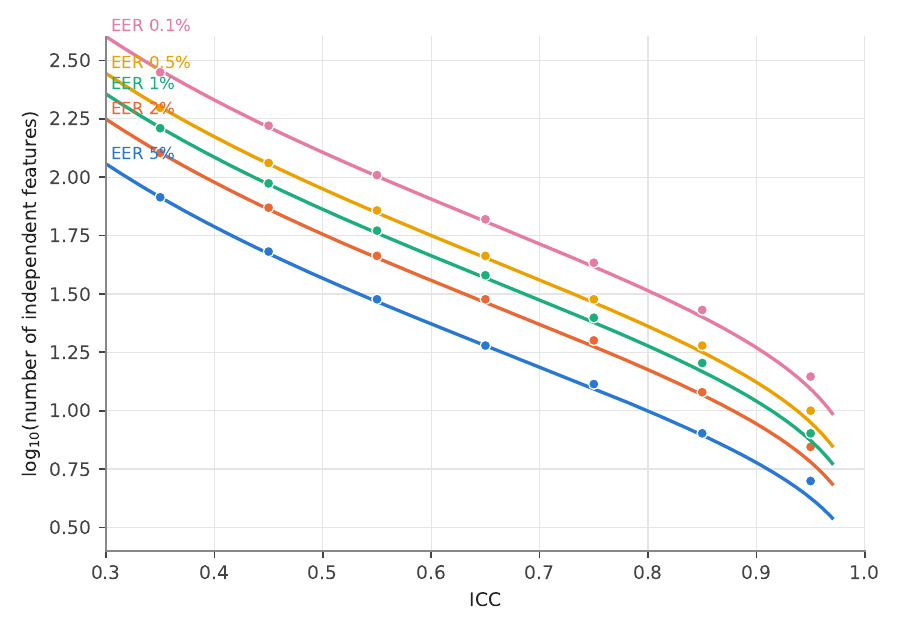}
\caption{Number of independent features required for five EER targets, on the logarithmic scale used in \cite{Friedman2019Linear}. Points: the counts found by simulation in that study. Lines: the inverse law (10) in its finite-$N$ form, $N=(2q/\atanh\rho)^2+1$. No parameter is fitted.}
\label{fig:loglinear}
\end{figure}

Table~\ref{tab:target} gives the cell-by-cell comparison summarized in Section~V-D of the main paper, and Fig.~\ref{fig:loglinear} shows it on the logarithmic scale of \cite{Friedman2019Linear}: the five near-parallel lines of that paper are the curves of (11). The intercepts quoted in \cite{Friedman2019Linear} ($2.59$ to $3.18$) are larger than $2\log_{10}(2q)$ ($1.0$ to $1.6$) because they are the values of straight lines extrapolated to $\ICC=0$, where $-2\log_{10}(\atanh\rho)$ diverges; the extrapolation adds about $1.6$ to each intercept.

\begin{table*}[t]
\centering
\caption{Independent features required to reach five EER targets: counts found by simulation in \cite{Friedman2019Linear} (first number), the finite-$N$ inverse law $N=(2q/\atanh\rho)^2+1$ rounded to the nearest integer (in parentheses), and the EER (\%) re-simulated here at the published count (in brackets).}
\label{tab:target}
\setlength{\tabcolsep}{6pt}
\begin{tabular}{@{}lccccc@{}}
\toprule
ICC & 5\% & 2\% & 1\% & 0.5\% & 0.1\% \\
\midrule
0.35 & 82 (82) [4.89] & 127 (127) [2.00] & 162 (163) [1.07] & 198 (200) [0.57] & 281 (287) [0.10] \\
0.45 & 48 (47) [4.83] & 74 (73) [1.95] & 94 (93) [0.97] & 115 (114) [0.49] & 166 (164) [0.13] \\
0.55 & 30 (29) [4.79] & 46 (45) [1.98] & 59 (58) [0.96] & 72 (70) [0.49] & 102 (101) [0.10] \\
0.65 & 19 (19) [5.00] & 30 (29) [1.94] & 38 (37) [1.06] & 46 (45) [0.44] & 66 (65) [0.09] \\
0.75 & 13 (12) [4.63] & 20 (19) [1.85] & 25 (24) [0.94] & 30 (29) [0.45] & 43 (41) [0.10] \\
0.85 & 8 (8) [4.83] & 12 (12) [2.14] & 16 (15) [0.77] & 19 (18) [0.44] & 27 (25) [0.09] \\
0.95 & 5 (4) [3.71] & 7 (6) [1.52] & 8 (7) [1.01] & 10 (9) [0.41] & 14 (12) [0.09] \\
\bottomrule
\end{tabular}
\end{table*}

\section{Unequal Persistence}
\label{sup:unequal}
Table~\ref{tab:unequal} gives the results discussed in Section~V-E of the main paper.

\begin{table}[t]
\centering
\caption{EER (\%) for $N=32$ features of unequal persistence (5,000 subjects): simulation versus the finite-$N$ law evaluated at the mean and at the median ICC.}
\label{tab:unequal}
\footnotesize
\setlength{\tabcolsep}{3pt}
\begin{tabular}{@{}lccrrr@{}}
\toprule
ICC distribution & mean & med. & sim. & law(mean) & law(med.) \\
\midrule
uniform 0.3--0.9 & 0.60 & 0.60 & 3.08 & 2.68 & 2.68 \\
uniform 0.1--0.5 & 0.30 & 0.30 & 19.4 & 19.4 & 19.4 \\
75\% at 0.3, 25\% at 0.9 & 0.45 & 0.30 & 9.12 & 8.86 & 19.4 \\
25\% at 0.3, 75\% at 0.9 & 0.75 & 0.90 & 0.70 & 0.34 & 0.002 \\
half 0.05, half 0.8 & 0.425 & 0.425 & 11.4 & 10.3 & 10.3 \\
same, $N=64$ & 0.425 & 0.425 & 4.58 & 3.59 & 3.59 \\
\bottomrule
\end{tabular}
\end{table}

\section{Persistence Varying Across Subjects}
\label{sup:subjects}
Table~\ref{tab:subjects} gives the results discussed in Section~V-F of the main paper.

\begin{table}[t]
\centering
\caption{EER (\%) for $N=32$ independent features when persistence varies across subjects (4,000 subjects): simulation (sim.), the finite-$N$ law at the mean persistence (law), the two-normal form (2-norm.), the mixture form (14) (mixt.), and the effective number of features $\Neff$ inferred from the simulated EER at the mean persistence. Normal rows: $\rho_i\sim\mathcal N(0.6,\mathrm{SD}^2)$; two-point rows: the stated fraction of subjects at $\rho=0.2$, the rest at $0.7$.}
\label{tab:subjects}
\footnotesize
\setlength{\tabcolsep}{3pt}
\begin{tabular}{@{}lccrrrrr@{}}
\toprule
$\rho_i$ & mean & SD & sim. & law & 2-norm. & mixt. & $\Neff$ \\
\midrule
normal & 0.60 & 0.00 & 2.59 & 2.68 & 2.60 & 2.68 & 32 \\
normal & 0.60 & 0.05 & 3.48 & 2.68 & 3.30 & 3.17 & 28 \\
normal & 0.60 & 0.10 & 4.70 & 2.75 & 4.92 & 4.53 & 25 \\
normal & 0.60 & 0.15 & 6.08 & 2.77 & 8.13 & 6.21 & 21 \\
normal & 0.60 & 0.19 & 8.43 & 2.79 & 11.66 & 8.00 & 17 \\
normal & 0.60 & 0.23 & 10.00 & 2.78 & 13.41 & 9.84 & 15 \\
5\% low & 0.68 & 0.10 & 3.45 & 1.09 & 1.87 & 3.54 & 20 \\
10\% low & 0.65 & 0.15 & 6.72 & 1.56 & 3.75 & 6.61 & 16 \\
20\% low & 0.60 & 0.20 & 10.32 & 2.59 & 6.85 & 10.59 & 14 \\
\bottomrule
\end{tabular}
\end{table}

\section{Non-Normal Features and the Range of the KCC Route}
\label{sup:kcc-range}
Kendall's $W$ is invariant to any monotone change of a feature, and cosine similarity is not. The KCC route (13) is therefore blind to exactly the departure from normality that changes the EER, and the useful question is how far from normal a feature set may be before that matters. We passed $N=32$ independent features with $\rho=0.6$ through $y=\operatorname{sign}(x)|x|^{p}$, which leaves every rank---and so every KCC---untouched while making the marginals heavier-tailed (function \texttt{exp11\_kcc\_bound.m}). At excess kurtosis $0$ the law in KCC units is exact ($2.68\%$ predicted, $2.69\%$ simulated). At excess kurtosis $2.9$ it is optimistic by $0.6$ EER points, at $8.8$ by $2.6$ points, and at $43$ by $6.9$ points, the prediction staying at $2.68\%$ throughout while the simulated EER rises to $9.6\%$. The mean Pearson $r$, which is not rank-based, tracks the simulation to within $0.6$ points up to an excess kurtosis of $43$. Two practical rules follow: read the law in KCC units only while the typical dimension has excess kurtosis of about $1$ or less, and, when it does not, replace each feature by the inverse-normal scores of its ranks before scoring---a monotone change that leaves the KCC unaltered and restored the law to within $0.02$ EER points in the heaviest-tailed case tested.

\section{Intercorrelated Features}
\label{sup:corr}
\begin{figure*}[t]
\centering
\includegraphics[width=0.9\textwidth]{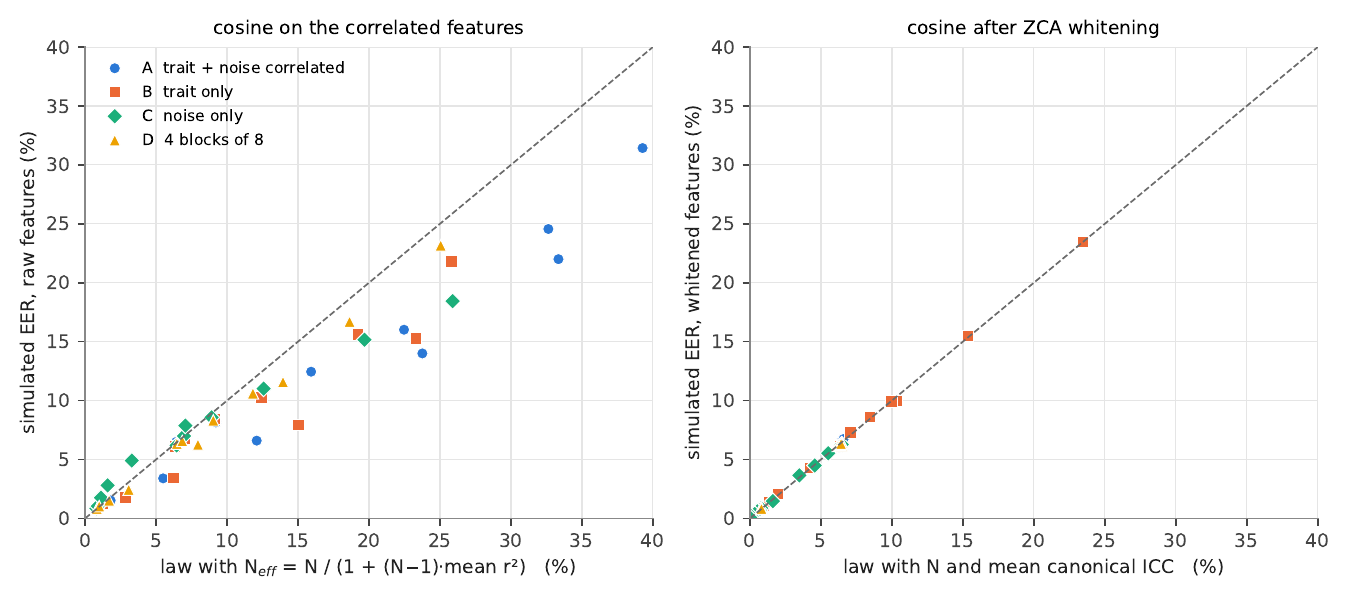}
\caption{Intercorrelated features, $N=32$, $\ICC\in\{0.5,0.7\}$, pairwise $r\in\{0,0.1,0.2,0.3,0.5,0.7\}$. Left: simulated EER of cosine scoring on the raw correlated features versus the law with $\Neff$ from (18). Right: the same features after ZCA whitening versus the law with $N$ and the mean canonical persistence \eqref{eq:canonical}.}
\label{fig:corr}
\end{figure*}

We imposed an exchangeable correlation $r$ (all pairs equal) on $N=32$ features with $\ICC=0.5$ or $0.7$ in four ways: (A) the same correlation in trait and noise, so that the within-session correlation is $r$ and the ICC of every feature is unchanged; (B) correlation only among the trait parts; (C) only among the noise parts; (D) four independent blocks of eight correlated features. In every case the per-feature ICC remains $\rho$, yet the EER of raw cosine scoring rises steeply with $r$---at $\ICC=0.7$ from $0.7\%$ at $r=0$ to $6.6\%$ at $r=0.3$ and $22\%$ at $r=0.7$ in case A. Note that $r$ is imposed on the trait or noise part, so the resulting within-session correlation is $r$ in case A, $\rho r$ in case B and $(1-\rho)r$ in case C; it is the measured within-session $\overline{r^2}$ that enters (18). Fig.~\ref{fig:corr} (left) plots the simulated EER against the law with $\Neff$ from (18). The rule tracks the simulation across all four structures: within $1.3$ points in the 24 conditions whose mean within-session $|r|$ is below $0.1$, within $3.5$ points up to $0.2$, and pessimistic by $4$--$11$ points at mean correlations of $0.3$--$0.7$, which arise only when a single common factor dominates (case A, and case B at $r\ge0.5$), because one very large eigenvalue of $\mathbf R$ depresses $\Neff$ more than it depresses the actual separability. Fig.~\ref{fig:corr} (right) shows the same features after ZCA whitening against \eqref{eq:canonical}: agreement within $0.4$ points in all 48 conditions. Whitening never worsened the EER by more than $1.7$ points and usually improved it substantially (case C, $\ICC=0.7$, $r=0.5$: $4.9\%$ raw, $0.2\%$ whitened), with the mean canonical persistence rising from $0.70$ to $0.80$ as shared noise was removed; in case B it fell to $0.55$, as shared trait information was removed, and the whitened EER of $4.3\%$ was still better than the raw $7.9\%$.

The impostor-based estimate (19) exceeded (18) by $5$--$25\%$ for mild or block-structured correlation and by up to a factor of $2.3$ when a single factor dominated (case A at $r\ge0.5$). Used with the genuine median in the equal-variance form of the law, it was within $1$ EER point in cases A, B and D for mean correlations up to $0.5$, but failed when the correlation sat in the session noise (case C), where the genuine Fisher-$z$ spread exceeded the impostor spread by up to $70\%$. The two-normal form with separate spreads, $\EER=\Phi\big(-(\mu_g-\mu_i)/(\sigma_g+\sigma_i)\big)$ in $z$ units, was within $1.2$ points in all 32 correlated conditions; the ratio $\sigma_g/\sigma_i$ is therefore a useful diagnostic, equal to one for independent features and above one when session noise is shared across features.

\section{Session-Level Changes}
\label{sup:session}
Table~\ref{tab:session} applies the session changes of Section~IV-E to session~2 of an $N=32$, $\ICC=0.7$ set. A shift of one standard deviation in every dimension raises the EER from $0.8\%$ to $6.4\%$; only $\ICC(2,1)$ registers it ($0.70\to0.47$), and the law evaluated with $\ICC(2,1)$ predicts $7.9\%$. A uniform rescaling by $3$ leaves the EER at $0.8\%$ but drives $\ICC(2,1)$ to $0.42$, so the law evaluated with it predicts $10.6\%$---far too pessimistic---while Pearson $r$ and KCC correctly report no change. The last two columns evaluate the law with $\hat\rho$ and $\widehat N_{\mathrm{eff}}$ read off the score distributions (Section~IV-E) and the two-normal form with separate spreads; both are within $0.4$ points in every row, whatever the session change.

\begin{table*}[t]
\centering
\caption{Cosine EER (\%) when session 2 is altered ($N=32$, $\ICC=0.7$, 3,000 subjects), the three persistence indices (means over the 32 features), the law evaluated with each index, the law evaluated with the score-distribution estimates $\hat\rho$ (genuine median) and $\widehat N_{\mathrm{eff}}$ (impostor spread), and the two-normal form with separate genuine and impostor spreads in Fisher-$z$ units.}
\label{tab:session}
\setlength{\tabcolsep}{0pt}
\begin{tabular*}{\textwidth}{@{\extracolsep{\fill}}lrrrrrrrr@{}}
\toprule
& & Pearson & & & & & law & two- \\
Change applied to session 2 & EER & $r$ & $\ICC(2,1)$ & KCC & law($r$) & law(ICC) & ($\hat\rho,\widehat N_{\mathrm{eff}}$) & normal \\
\midrule
none & 0.80 & 0.70 & 0.70 & 0.84 & 0.80 & 0.80 & 0.74 & 0.78 \\
shift $+1$ SD, every dimension & 6.40 & 0.70 & 0.47 & 0.84 & 0.79 & 7.93 & 6.55 & 6.47 \\
shift random per dimension (SD 1) & 8.10 & 0.70 & 0.48 & 0.84 & 0.80 & 7.11 & 8.05 & 7.74 \\
uniform rescale $\times3$ & 0.84 & 0.70 & 0.42 & 0.84 & 0.78 & 10.56 & 0.73 & 0.76 \\
random rescale per dimension & 2.73 & 0.71 & 0.60 & 0.84 & 0.72 & 2.67 & 2.60 & 2.34 \\
\bottomrule
\end{tabular*}
\end{table*}

\section{Score Distributions of the Embeddings}
\label{sup:stage1}
\begin{figure*}[t]
\centering
\includegraphics[width=0.9\textwidth]{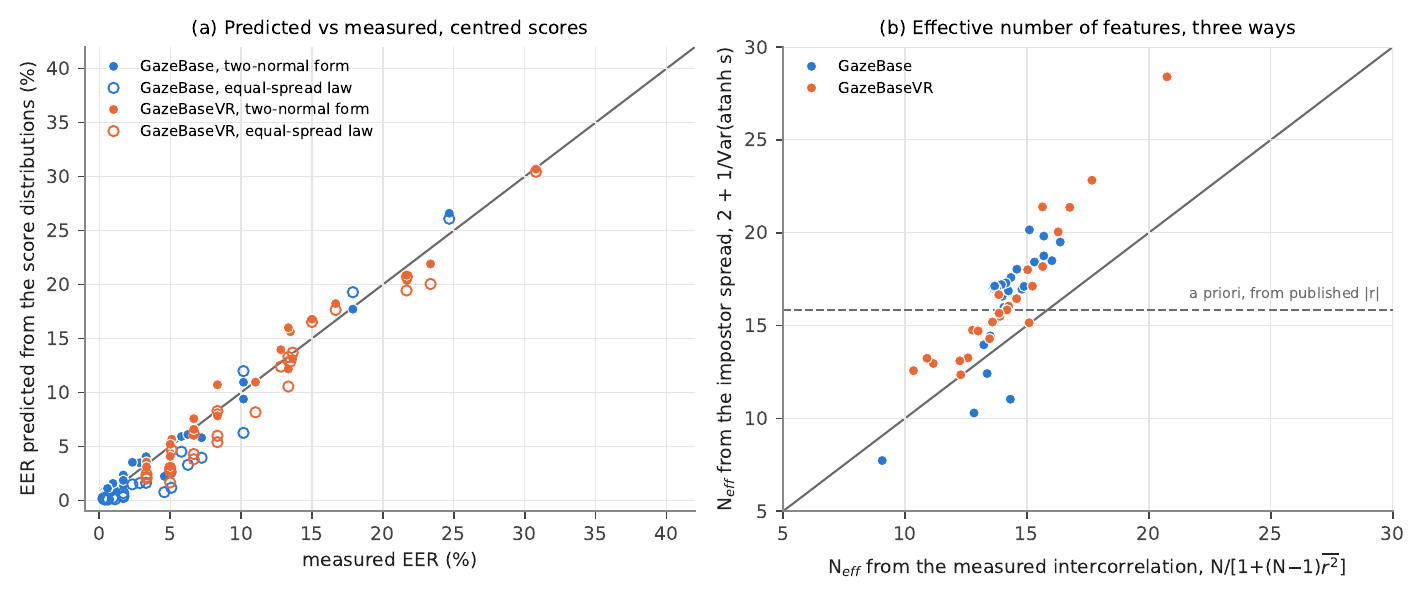}
\caption{Score distributions of the stored Eye Know You Too embeddings, 56 of the 61 conditions of \cite{Raju2026TBIOM}, each centred on the mean of its own templates. (a) EER predicted from the genuine and impostor scores---equal-spread law with $\hat\rho$ and $\widehat N_{\mathrm{eff}}$ (open) and two-normal form (filled)---against the EER measured on the same scores; the line is identity. (b) Effective number of independent features from the impostor spread (19) against the value from the measured intercorrelation (18); dashed: the a-priori value $16$ from the published intercorrelation.}
\label{fig:stage1}
\end{figure*}

Fig.~\ref{fig:stage1} shows the results of Section~VI-C of the main paper.

\section{Which Index Predicts the Score Centre}
\label{sup:index}
Stage~2 of the protocol asks which per-dimension persistence index predicts $\hat\rho$, the centre of the genuine score distribution. Table~\ref{tab:indexfit} answers it for the 56 conditions. All three indices track $\hat\rho$ closely across the whole range, and to within rounding they do so equally well (correlations $0.992$ to $0.993$). These embeddings are also well inside the range in which the KCC may stand in for the correlation (Section~V-G): the median dimension has excess kurtosis $0.25$, and $8\%$ of dimensions exceed $2$. The ICC and the Pearson $r$ agree with each other to within $0.02$ (median $0.005$), so these manipulations introduce no appreciable session effect---unsurprisingly, since each was applied to both sessions alike---and the simulations of Section~V-I therefore do not separate the indices here.

All three, however, understate $\hat\rho$: by $0.030$ for the Pearson $r$, $0.036$ for the ICC and $0.043$ for the KCC. The reason is in the derivation: the cosine weights each dimension by its variance, so the persistence that enters it is the variance-weighted mean of the per-dimension persistences, not the plain mean. Weighting by $\sqrt{\Var(x_{k1})\Var(x_{k2})}$ reduces the bias to $-0.013$ and is the closest of the four (correlation $0.994$, largest departure $0.093$). The practical rule is that a mean index predicts the score centre up to a small positive offset, and the variance-weighted index predicts it with the smallest bias and the smallest spread.

\begin{table}[t]
\centering
\caption{How well each per-dimension persistence index predicts $\hat\rho$, the median genuine cosine score, over the 56 embedding conditions (centred templates). Offsets are $\hat\rho$ minus the index.}
\label{tab:indexfit}
\footnotesize
\setlength{\tabcolsep}{4pt}
\begin{tabular}{@{}lccc@{}}
\toprule
Index over the 128 dimensions & corr. & mean off. & max $|$off.$|$ \\
\midrule
mean Pearson $r$ & 0.992 & $+0.030$ & 0.093 \\
mean $\ICC(2,1)$ & 0.993 & $+0.036$ & 0.094 \\
KCC, mapped by (12) & 0.993 & $+0.043$ & 0.107 \\
variance-weighted Pearson $r$ & \textbf{0.994} & $\mathbf{-0.013}$ & \textbf{0.093} \\
\bottomrule
\end{tabular}
\end{table}

This also settles why the GazeBaseVR conditions fall below the curve in Fig.~4. It is not the index: the KCC understates $\hat\rho$ by $0.031$ on the decimation conditions and $0.045$ on the short recordings, the same offset as everywhere else. It is $\Neff$, which those manipulations move away from the fixed value of $16$ that the curve assumes. Shortening the recording raises it, and both estimates agree that it does: with one sequence averaged instead of eight, the impostor spread gives $28$ and the intercorrelation $21$, against $16$ and $14$ with eight---averaging fewer sequences leaves the dimensions sharing less session-level variation. Decimation raises the impostor-based estimate to $21$ at both $25$ and $10$\,Hz while the intercorrelation-based one stays near $16$, the regime in which (18) was pessimistic in simulation and (19) the more accurate of the two (Section~V-H). In both cases a curve drawn at a single $\Neff$ must be pessimistic, and the score-based form of Section~VI-C, which reads $\hat\rho$ and $\widehat N_{\mathrm{eff}}$ off the same scores, predicts these conditions within one EER point.

\section{Keystroke Dynamics: Further Readings}
\label{sup:keystroke}
Three further readings support the treatment developed above. First, $\Neff$: the 31 features behave like $6.9$ independent ones by the impostor spread and $3.5$ by the intercorrelation, whose mean squared value is $0.26$---far higher than for the embeddings---so a hand-crafted feature set again delivers a handful of effective features, as the 29 subsets of Section~VI-A did. Second, non-normality: the raw latencies have a mean excess kurtosis of $6.6$, outside the range in which the rank-based index may be substituted (Section~V-G); replacing each feature by the inverse-normal scores of its ranks brings the excess kurtosis to $-0.2$, and the law continues to hold ($8.27\%$ predicted against $7.96\%$ measured at one day). Third, averaged enrollment: as the number of repetitions averaged into each template grows from 1 to 50, the measured persistence rises from $0.47$ to $0.87$ while Spearman--Brown predicts $0.98$, because repetitions within a session share session-level variation rather than being independent---the saturation anticipated in Section~IV-F. Fed the measured centroid persistence instead, the law tracks the EER from $25.5\%$ at one repetition to $5.9\%$ at fifty.

\section{Baseline Conditions of Each Modality}
\label{sup:baseline}
Table~\ref{tab:baseline} gives the baseline comparison discussed in Section~VI-I of the main paper.

\begin{table}[t]
\centering
\caption{Measured EER (\%) of the baseline condition of each modality and the predictions of the law with one persistence ($\tau=0$), the two-normal form and the general form with the measured spread (14); impostor route, $x=\tau\sqrt{\Neff-2}$. Keystroke: raw or rank-normalized features, probe in session 2 (S2) or 8 (S8). Faces: held-out halves at 512 dimensions (two-normal form not evaluated). The $\tau=0$ column reads $m$ from the mean genuine Fisher-$z$ score; the equal-spread values of Sections~VI-F and~VI-G use the median genuine score instead.}
\label{tab:baseline}
\footnotesize
\setlength{\tabcolsep}{3.5pt}
\begin{tabular}{@{}lrrrrr@{}}
\toprule
Condition & $x$ & measured & $\tau=0$ & two-norm. & general \\
\midrule
GazeBase, unmanip. & 1.10 & 0.44 & 0.21 & 1.07 & 1.03 \\
GazeBaseVR, unmanip. & 0.68 & 3.33 & 2.29 & 3.52 & 3.47 \\
Keystroke raw, S2 & 0.00 & 5.88 & 7.16 & 5.87 & 7.16 \\
Keystroke raw, S8 & 0.00 & 17.29 & 19.59 & 18.33 & 19.59 \\
Keystroke rank, S2 & 0.36 & 7.96 & 7.62 & 8.27 & 8.24 \\
Keystroke rank, S8 & 0.45 & 18.71 & 20.05 & 21.16 & 21.16 \\
ECG & 1.88 & 14.63 & 3.92 & 13.05 & 13.75 \\
Faces 3--5 years & 2.93 & 3.03 & 0.00 & --- & 2.69 \\
Faces 6--10 years & 3.06 & 3.68 & 0.00 & --- & 3.32 \\
Faces 11--20 years & 3.00 & 5.66 & 0.01 & --- & 5.04 \\
Faces 21--30 years & 2.72 & 6.05 & 0.02 & --- & 4.93 \\
Faces 31+ years & 2.43 & 6.34 & 0.07 & --- & 5.35 \\
\bottomrule
\end{tabular}
\end{table}

\section{Design Curves}
\label{sup:design}
\begin{figure*}[t]
\centering
\includegraphics[width=0.9\textwidth]{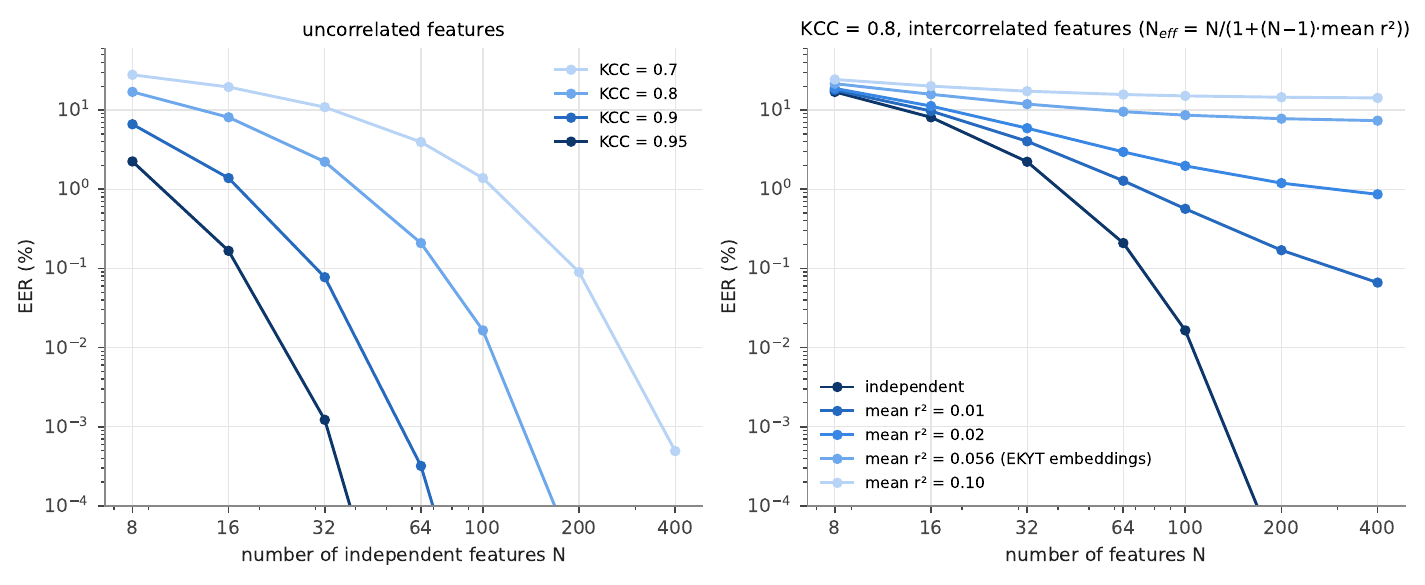}
\caption{Left: EER versus the number of independent features for four levels of KCC. Right: at $\KCC=0.8$, the same relation as the mean squared intercorrelation grows; at $\overline{r^2}=0.056$, the value for the embeddings of \cite{Raju2026TBIOM}, the EER flattens near $8\%$ because $\Neff$ saturates at about $18$.}
\label{fig:design}
\end{figure*}

Fig.~\ref{fig:design} shows graphically the relations of Table~V of the main paper.

\section{Which Form of the Law to Use}
\label{sup:which}
Table~\ref{tab:which} collects the cases met in the main paper (Section~VII).

\begin{table}[t]
\centering
\caption{Which form of the law to use. $\sigma_g/\sigma_i$: ratio of the genuine to the impostor Fisher-$z$ standard deviation, read from a pilot set.}
\label{tab:which}
\footnotesize
\setlength{\tabcolsep}{3pt}
\begin{tabular}{@{}p{0.27\columnwidth}p{0.36\columnwidth}p{0.31\columnwidth}@{}}
\toprule
Diagnostic & Form & Example \\
\midrule
$\sigma_g/\sigma_i<1$ (heavy-tailed features) & rank-normalize first (Section~V-G); otherwise the two-normal form & raw keystroke latencies ($0.87$--$0.90$) \\
$\sigma_g/\sigma_i\approx1$ ($\tau\approx0$) & the law (8) at $\hat\rho$ and $\Neff$ & rank-normalized keystroke ($1.06$--$1.10$) \\
$\sigma_g/\sigma_i>1$; normal and measured spread agree & closed form (15), equivalently the two-normal form & eye-movement embeddings ($1.3$) \\
$\sigma_g/\sigma_i>1$; they disagree (heavy lower tail) & mixture (14): $\rho_i$ from three or more sessions if the spread is between people, deconvolved pilot genuine scores if between recordings & ECG ($2.1$), faces ($2.6$--$3.2$) \\
\bottomrule
\end{tabular}
\end{table}

\section{A Staged Validation Protocol}
\label{sup:protocol}
This section gives the full protocol summarized in Section~VIII of the main paper. Stages~1--3 have been carried out; Stages~4 and~5 have not.

\emph{Stage 1: does the two-normal Fisher-$z$ picture hold for embeddings?} (Carried out in Section~VI-C: after centring, yes, in the two-normal form.) For each condition of \cite{Raju2026TBIOM} whose stored embeddings reproduce its published EER (56 of the 61; Section~VI-C), compute the genuine and impostor cosine scores of the centroids, $\hat\rho$ (median genuine score), $\widehat N_{\mathrm{eff}}$ from (19), and the predicted $\EER=\Phi(-\tfrac12\sqrt{\widehat N_{\mathrm{eff}}-1}\,\atanh\hat\rho)$. Compare with the measured EER on the $\dprime$ scale, $\dprime=2\,\Phi^{-1}(1-\EER)$, where the law predicts a line of slope one through the origin. Check also that $\Var(\atanh s_{\mathrm{gen}})\approx\Var(\atanh s_{\mathrm{imp}})$; if the genuine spread is larger, the two-normal form $\EER=\Phi\big(-(\mu_g-\mu_i)/(\sigma_g+\sigma_i)\big)$, with $\mu$ and $\sigma$ the means and standard deviations of the genuine ($g$) and impostor ($i$) scores in $z$ units, is the appropriate, still parameter-free, comparison (it was within $1.2$ points in every simulated condition of Sections~V-H and~V-I). With 59 subjects the EER is quantized in steps of $1/59$ on the genuine side, so EER values below about $1\%$ cannot be tested, but the score-distribution moments are well estimated (59 genuine and 3,422 impostor scores); bootstrap over subjects gives confidence intervals, and a synthetic set with 59 subjects calibrates the scatter to expect.

\emph{Stage 2: which persistence index predicts $\hat\rho$?} (Carried out in Section~VI-D: all three do, equally well to within rounding, and the offset common to all three is explained by the variance weighting of the cosine.) Per condition, compute over the 128 dimensions the mean (and median) Pearson $r$, $\ICC(2,1)$ and KCC of the centroids and regress $\hat\rho$ on each. Table~\ref{tab:session} predicts agreement when there is no session change and identifies the kind of change from the pattern of disagreement. For the GazeBaseVR decimation and short-recording conditions that fall below the curve in Fig.~4, Section~VI-D finds that the index is not at fault: $\Neff$ itself rises under those manipulations.

\emph{Stage 3: does intercorrelation set $\Neff$?} (Carried out in Section~VI-E: yes, and the EER floor that remains is set by persistence varying across comparisons.) Compare $\widehat N_{\mathrm{eff}}$ from Stage~1 with (18) computed from the within-session correlation matrix of the centred embeddings. Then draw random subsets of $8,16,32,64$ and $128$ dimensions, recompute EER and $\widehat N_{\mathrm{eff}}$ for each, and check that $\widehat N_{\mathrm{eff}}(N)$ follows $N/[1+(N-1)\overline{r^2}]$ and saturates near $1/\overline{r^2}\approx18$. This is the embedding analogue of the feature-count experiment of \cite{Friedman2019Linear} and needs no new data.

\emph{Stage 4: does whitening deliver the canonical prediction?} Estimate the whitening matrix on training-round embeddings with shrinkage \cite{Ledoit2004} (128 dimensions require many more subjects than 59), apply it to the test embeddings, compute the canonical spectrum, and check $\EER=\Phi(-\tfrac12\sqrt{128}\,\atanh\bar\lambda)$. The gain over raw cosine scoring is the practical payoff; a spectrum dominated by a few components identifies decorrelation in training as the next lever.

\emph{Stage 5: close the loop on the 2017 datasets.} Recompute the 29 subsets with mean ICC, with the ICC of the retained PCA components, and with the participation ratio $(\sum_k\sigma_k^2)^2/\sum_k\sigma_k^4$ of the component variances; the law says that the implied $\Neff$ of Fig.~3 should equal that ratio and $\hat\rho$ the variance-weighted component persistence. A remaining shortfall of the implied $\Neff$ below the ratio points to persistence that varies across subjects; the excess of the genuine over the impostor Fisher-$z$ variance estimates that spread directly (Section~IV-B), and the mixture form (14) should then close the gap.

\section{Code}
\label{sup:code}
All results were produced by MATLAB-syntax scripts executed under GNU Octave~8.4 or MATLAB R2024a without toolboxes: \texttt{eerLaw.m} (8), \texttt{nFeaturesLaw.m} (10), \texttt{eerLawKCC.m}, \texttt{kcc2rho.m}, \texttt{rho2kcc.m} (12)--(13), \texttt{eerExact.m} \eqref{eq:fisher-density}, \texttt{eerLawMixture.m} (14), \texttt{unifiedLaw.m} (15), \texttt{synthFeatures.m}, \texttt{cosineEER.m}, \texttt{kcc2.m}, and the experiment scripts \texttt{exp1}--\texttt{exp17} that generate every table and figure, with \texttt{exp8\_summary.m} printing the statistics quoted in Sections~VI-C and~VI-D. The keystroke and ECG data of Sections~VI-F and~VI-G will be included in the package. The face embeddings are computed from the images by the Python scripts \texttt{face\_embed\_lfw.py} and \texttt{face\_embed\_agedb.py}. The InsightFace models are released for non-commercial research only, and the AgeDB and LFW images may not be redistributed; the code package contains the scripts that regenerate the embeddings. The package will be made available upon acceptance.

\bibliographystyle{unsrt}
\bibliography{ms}